\documentclass[sigconf]{acmart}

\AtBeginDocument{%
  }

\usepackage{amsmath}
\usepackage{mathtools}
\usepackage{amsthm}

\usepackage{pifont}
\usepackage{graphicx}
\usepackage{multirow}
\usepackage{subfigure}
\usepackage{booktabs}
\usepackage{enumitem}
\usepackage{color}
\usepackage{xcolor}

\newcommand{\proposedmodel}{CODA}
\newcommand{\panelxy}[3]{%
  \raisebox{#2}{\makebox[\linewidth]{\hspace*{#1}\footnotesize\textbf{(#3)}}}%
}

\renewcommand\footnotetextcopyrightpermission[1]{} 
\begin{document}

\title{Score-based Conditional Out-of-Distribution Augmentation for Graph Covariate Shift}


\author{Bohan Wang}
\affiliation{%
  \institution{Emory Unversity}
  \country{USA}
  }
\email{bohan.wang2@emory.edu}

\author{Yurui Chang}
\affiliation{%
  \institution{The Pennsylvania State University}
  \country{USA}
  }
\email{yjc5487@psu.edu}

\author{Wei Jin}
\affiliation{%
  \institution{Emory Unversity}
  \country{USA}
  }
\email{wei.jin@emory.edu}

\author{Lu Lin}
\affiliation{%
  \institution{The Pennsylvania State University}
  \country{USA}
  }
\email{lxl5598@psu.edu}

\begin{abstract}
Distribution shifts between training and testing datasets significantly impair the model performance on graph learning. A commonly-taken causal view in graph invariant learning suggests that stable predictive features of graphs are causally associated with labels, whereas varying environmental features lead to distribution shifts. In particular, covariate shifts caused by unseen environments in test graphs underscore the critical need for out-of-distribution (OOD) generalization. Existing graph augmentation methods designed to address the covariate shift often  disentangle the stable and environmental features in the input space, and selectively perturb or mixup the environmental features. However, such perturbation-based methods heavily rely on an accurate separation of stable and environmental features, and their exploration ability is confined to existing environmental features in the training distribution. To overcome these limitations, we introduce a novel distributional augmentation approach enabled by a tailored score-based conditional graph generation strategies to explore and synthesize unseen environments while preserving the validity and stable features of overall graph patterns. Our comprehensive empirical evaluations demonstrate the enhanced effectiveness of our method in improving graph OOD generalization.

\end{abstract}

\ccsdesc[500]{Computing methodologies~Neural networks}
\ccsdesc[500]{Mathematics of computing~Graph algorithms}



\maketitle

\pagestyle{plain} 

\section{Introduction}

Deep learning has become the dominant paradigm for analyzing graph-structured data across diverse domains. However, most graph learning methods assume that training and test graphs are independently and identically distributed (i.i.d.)., an assumption that frequently fail in real-world setting~\cite{gui2022good, huang2024enhancing}. On the Web, spanning hyperlink graphs, social networks, recommendation and e-commerce interaction graphs, and knowledge graphs, distribution shift frequently arise due to product launches, policy or UI changes, seasonality, bot activity, and rapid community growth. Prior studies have shown that such shifts can lead to substantial performance degradation~\cite{gui2022good, sui2024unleashing}, motivating growing interest in graph out-of-distribution (OOD) generalization, including invariant graph learning~\cite{chen2022learning, wu2022discovering, huang2024enhancing} and graph data augmentation~\cite{rong2019dropedge, wang2021mixup, han2022g, yao2022improving, sui2024unleashing, ligraph}.

Following~\cite{gui2022good, sui2024unleashing}, we distinguish \emph{correlation shift}, where environment–label correlations differ but test environments are represented during training, from \emph{covariate shift}, where the test environments are not present in the training dataset. A causal view in invariant graph learning~\cite{chen2022learning, wu2022discovering} posits that label-determining \emph{stable} patterns are invariant across environments, while \emph{environmental} patterns vary. Prior work largely targets correlation shift by isolating stable patterns~\cite{miao2022interpretable, chen2022learning, wu2022discovering}. We focus instead on the underexplored \emph{covariate} shift in graph learning, which is pervasive in Web settings (e.g., emerging communities or new interaction regimes unseen at train time).

Existing graph augmentation techniques typically perturb graph structures or features by mixing or dropping edges~\cite{rong2019dropedge, han2022g, sui2024unleashing}. While effective with regularized contexts, such operations offer limited exploration ability to  unseen environments and risk generating semantically inconsistent or unreliable samples (e.g., invalid molecules in chemistry), as indiscriminate edits may distort label-relevant stable patterns, yielding \emph{uncontrolled} augmented distributions. Controlled methods~\cite{sui2024unleashing} help but often require an explicit, accurate split of stable vs.\ environmental subgraphs—nontrivial and sometimes infeasible on Web graphs. This raises a key question: 

\begin{center}
\emph{Can we explore beyond the training environments in a controllable way while preserving label-critical stable patterns?}
\end{center}

We provide an affirmative answer via a score-based \emph{conditional} graph generation strategy for covariate-shift mitigation. We formulate OOD augmentation as generating graphs conditioned on a target label and an \emph{exploration parameter} that controls the degree of distributional deviation, building on the widely used graph-generation hypothesis~\cite{wu2022discovering, yang2022learning, gui2022good, chen2022learning}. Inspired by diffusion models~\cite{song2020score}, we define a forward corruption process and a novel reverse denoising process that jointly (i) retain predictive stable patterns and (ii) inject environment exploration. Our conditional score function incorporates both label information and the degree of OOD-ness, enabling \emph{\underline{C}onditional \underline{O}ut-of-Distribution \underline{D}iffusion \underline{A}ugmentation} (\proposedmodel) to generate graphs likely to preserve stable patterns while probing \emph{underrepresented (low-density) regions}\footnote{Low-density regions are parts of the class-conditional data manifold with few samples; see Section~\ref{secsec:prior_knowledge}.} of the class-conditioned manifold beyond the training graph space.

\proposedmodel{} inherits diffusion-model robustness, helping ensure validity (preventing, e.g., non-viable molecules) without explicitly separating stable from environmental subgraphs. We evaluate on synthetic and real-world graph classification tasks under diverse covariate shifts, which are commonly observed in \emph{Web contexts}.  
For instance, motif-frequency variations emulate evolving community or interaction structures; changes in graph size correspond to thread-depth or document-length drift; and feature/structural shifts reflect the dynamics of user/item attributes and connectivity patterns in online platforms. Across these settings, \proposedmodel{} achieves consistent gains over state-of-the-art invariant-learning and augmentation methods, enabling controlled exploration of OOD environments while preserving stable patterns.

Our main contributions are summarized as follows:
\begin{itemize}[leftmargin=*]
\item We propose a novel conditional graph generation-based  augmentation framework that addresses covariate distribution shifts. Our method enables \emph{controlled} exploration of environmental patterns while preserving stable patterns, without requiring explicit subgraph decomposition.
\item We develop a unified generator for OOD \emph{structures}, \emph{node features}, and \emph{edge features}, handling feature, structural, and joint shifts which are pervasive in Web contexts.
\item Extensive experiments across synthetic, semi-artificial, molecular, and Web-analogous text/review graph datasets demonstrate consistent gains over strong OOD generalization baselines.
\end{itemize}

\section{Related Work}


Graph-structured data, such as hyperlink graphs, social networks and e-commerce interaction graphs in Web, are inherently complex, characterized by the intricate challenges of irregularity and nuanced structural information. This complexity gives rise to graph OOD problems that not only necessitate addressing shifts in feature distributions but also demand attention to variations of structural distributions.  In this context, we summarize two principal categories of algorithms for graph OOD robustness: (i) \emph{invariant graph learning} strategies, which aim to ensure model stability across varying distributions; and (ii) \emph{graph data augmentation} techniques, designed to enhance model generalizability by simulating diverse distribution scenarios.

\noindent\textbf{Invariant Graph Learning. }
The concept of invariant graph learning, inspired from seminal works~\cite{arjovsky2019invariant,  rosenfeld2020risks, ahuja2021invariance}, targets on identifying stable graph structures (e.g., subgraphs) or representations (predictors) that remain consistent across different environments, thereby enhancing OOD generalization. This is achieved by capturing salient graph features and minimizing empirical risks across varying conditions. In scenarios where establishing causality is complex or where strong assumptions may not hold, the task can be approximated by identifying features that demonstrate invariance under distributional shifts, thereby facilitating OOD generalization~\cite{li2022out}. Effective OOD generalization is achieved by basing predictions solely on invariant information~\cite{li2022out}. For example, DIR~\cite{wu2022discovering} distinguishes between invariant and environment-specific subgraphs by creating varied interventional distributions on the training distribution. CIGA~\cite{chen2022learning} further explores this domain by employing synthetic environments and the graph generation process to identify stable features under various distribution shifts. However, this line of research assumes \emph{access to test environments during training}, which is an unrealistic assumption given the impracticality of covering all possible test scenarios. Training in limited environments reduces spurious correlations but fails to generalize to new, unseen environments. DISGEN~\cite{huang2024enhancing} gains promising results in disentangling the size factors from graph representations by minimizing the shared information between size- and task-related information, however, the technique is constrained to handle size generalization. Recently, CANET~\cite{wu2024graph} introduces an environment estimator that infers a pseudo-environment, enabling the model to capture stable features under node-level distribution shifts 
without requiring prior knowledge of environment labels.
Nevertheless, CANET does not address OOD issues in other applications, such as molecular graphs. In this work, we propose a framework capable of generalizing to unseen environments characterized by differences not only in graph size but also in graph structure, node features, and edge features.

\noindent\textbf{Graph Data Augmentation.}
Beyond invariant graph learning, graph data augmentation aims to diversify the training distribution, thereby enhancing the OOD generalization of models. DropEdge~\cite{rong2019dropedge} introduces randomness by selectively removing edges, thus varying the training data's structure. M-Mixup~\cite{wang2021mixup} enriches the dataset by interpolating diverse and irregular graphs within semantic space. $\mathcal{G}$-Mixup~\cite{han2022g} extends mixup to graph classification by interpolating across graph generators (graphons). Adversarial augmentation techniques, such as FLAG~\cite{kong2022robust}, perturb node features via gradients, while AIA~\cite{sui2024unleashing} learns adversarial masks to probe environmental discrepancies. Despite these advances, methods that edit only the observed training graphs often have limited ability to explore truly novel environments and can distort label-determining patterns, yielding \emph{uncontrolled} augmentations. Environment-aware frameworks~\cite{ligraph} partially address this by using environment metadata to linearly explore structures and features; however, they rely on high-quality, diverse environment labels that are costly and often infeasible to obtain. GRATIN~\cite{abbahaddou2025graph} takes a different route: it fits a Gaussian mixture on learned graph embeddings and samples synthetic representations—supported by a Rademacher-complexity analysis—to achieve efficient gains in generalization without editing raw graphs. In contrast, our work introduces a \emph{generation-based} augmentation method that dispenses with environment labels and generated graphs end-to-end (structures, node features, and edge features), enabling \emph{controlled} exploration beyond the training graph space while preserving stable, label-determining patterns.

\vspace{-10pt}
\section{Problem Formulation}
\label{sec:problem_defination}

\textbf{Notations.}
We represent a graph with $n$ nodes as $G=(\boldsymbol{A}, \boldsymbol{X}, \boldsymbol{E})$, where $\boldsymbol{A} \in \mathbb{R}^{n \times n}$ is the adjacency matrix, $\boldsymbol{X} \in \mathbb{R}^{n \times a}$ denotes $a$-dimensional node features and $\boldsymbol{E} \in \mathbb{R}^{n \times n \times b}$ encodes $b$-dimensional edge features. 
Without the loss of generality, we focus on the graph classification task where each graph $G$ is associated with a label $Y \in \mathcal{Y}$, determined by a predefined labelling rule $\mathcal{G} \rightarrow \mathcal{Y}$. 
Following invariant learning~\cite{ahuja2021invariance, chen2022learning}, we denote the graph dataset as $\mathcal{D} = \{(G_i^e, Y_i^e)\}_{e\in\mathcal{E}_{\mathrm{all}}}$, where $(G_i^e, Y_i^e)\sim P_e(G, Y)$ is an i.i.d. draw in the environment $e$ sampled from all possible environments $\mathcal{E}_{\mathrm{all}}$. The complete dataset can be partitioned into a training set $\mathcal{D}_{\mathrm{tr}} = \{(G_i^e, Y_i^e)\}_{e \in \mathcal{E}_{\mathrm{tr}}}$ and a test set $\mathcal{D}_{\mathrm{te}} = \{(G_i^e, Y_i^e)\}_{e \in \mathcal{E}_{\mathrm{te}}}$, where $\mathcal{E}_{\mathrm{tr}}$ and $\mathcal{E}_{\mathrm{te}}$ index the training and testing environments, respectively. In practice, environment information may not be explicitly given, and we further denote the training distribution as $P_{\mathrm{tr}}(G, Y)$ and the testing distribution as $P_{\mathrm{te}}(G, Y)$.

\subsection{Graph Covariate Shift}
\label{sec:types_covariate_shifts}
We study OOD graph classification under \emph{covariate shift}: the label mechanism is stable while the input distribution shifts, i.e.,
\[
P_{\mathrm{tr}}(Y\mid G)=P_{\mathrm{te}}(Y\mid G)
\quad\text{but}\quad
P_{\mathrm{tr}}(G)\neq P_{\mathrm{te}}(G).
\]
This mismatch arises from limited training coverage and changing test conditions. Prior work~\cite{ligraph} distinguishes: (i) \textbf{feature shift} $P_{\mathrm{tr}}(\mathbf{X})\neq P_{\mathrm{te}}(\mathbf{X})$ with $P_{\mathrm{tr}}(\mathbf{A},\mathbf{E})=P_{\mathrm{te}}(\mathbf{A},\mathbf{E})$ (e.g., GOOD-CMNIST~\cite{gui2022good}, where color defines environments); (ii) \textbf{structural shift} $P_{\mathrm{tr}}(\mathbf{A},\mathbf{E})\neq P_{\mathrm{te}}(\mathbf{A},\mathbf{E})$; and (iii) \textbf{joint shift} $P_{\mathrm{tr}}(\mathbf{A},\mathbf{X},\mathbf{E})\neq P_{\mathrm{te}}(\mathbf{A},\mathbf{X},\mathbf{E})$ (e.g., GOOD-Motif size/basis changes~\cite{gui2022good}, or scaffold splits in GOOD-HIV~\cite{gui2022good}). All three forms challenge OOD generalization; our framework is designed to address them.

\subsection{Graph Classification under Covariate Shift}
With only observing the training set $\mathcal{D}_{\mathrm{tr}}$ sampled from the training distribution $P_{\mathrm{tr}}$ in training environments $\mathcal{E}_{\mathrm{tr}}$, our generalization objective under graph covariate shift is to train an optimal graph classifier $f: \mathcal{G}\rightarrow \mathcal{Y}$ that performs well across any possible environments $\mathcal{E}_{\mathrm{all}} \supseteq \mathcal{E}_{\mathrm{tr}}$. We formulate this  goal as the following minimization problem:

\begin{equation}
    \underset{f}{\min} \  \mathbb{E}_{e\in \mathcal{E}_{\mathrm{all}}} \mathbb{E}_{(G^e,Y^e) \sim P_e(G, Y)}[\ell(f(G^e), Y^e)],
    \label{eq:ood}
\end{equation}
where $\ell(\cdot, \cdot)$ denotes the loss function for graph classification and the expectation is with respect to graphs under all possible environments. However in practice, the training environments $\mathcal{E}_{\mathrm{tr}}$ may not cover all environments, causing degraded classification performance when applying the learned classifier in unseen test environments. This covariate shift calls for an effective manner to sufficiently explore unseen data distribution or environments during model training. We summarize and discuss the various types of graph covariate shifts in detail in Appendix.

\subsection{Issues in Environmental Augmentation}
To augment the training distribution for mitigating graph covariate shift, existing solutions often approach Eq.~(\ref{eq:ood}) 
by separating and augmenting the environments: 
\begin{equation}
    {\min_f} \  \mathbb{E}_{e\in \{\mathcal{E}_{\mathrm{tr}}\cup \mathcal{E}_{\mathrm{aug}}\}} \mathbb{E}_{(G^e,Y^e) \sim P_e(G, Y)}[\ell(f(G^e), Y^e)],
\end{equation}
where the augmented environments $\mathcal{E}_{\mathrm{aug}}$ are obtained 
based on either interpolating explicitly given environmental labels~\cite{ligraph} or perturbing implicitly separated environmental components~\cite{sui2024unleashing, wu2022discovering, miao2022interpretable, chen2022learning}. Obtaining accurate environmental labels and components itself could be high-cost and nontrivial tasks. Furthermore, without explicit environmental information and additional assumptions, separating environmental components could be inherently unfeasible~\cite{chen2024does}, which limit the practicability of such strategy. Additionally, the subgraph perturbations based augmentation is mainly operated by edge dropping~\cite{sui2024unleashing, rong2019dropedge} and mixup~\cite{han2022g}, which is confined to existing subgraphs in training data and could cause invalid samples (e.g., generating molecules that are chemically invalid).

\section{Score-based Conditional Out-of-Distribution Diffusion Augmentation (\proposedmodel)}
\label{sec:methodology}


In this section we present \proposedmodel, a score-based augmentation framework that generates graph samples exploring new environments while preserving label-determining structure. We first introduce a distribution-level view of augmentation, then describe the forward and reverse diffusion processes, and finally formalize the working principles of \proposedmodel.

\noindent\textbf{Our New Perspective: Distributional Augmentation with OOD Control.}
To overcome the limitations of existing strategies, this work formulates the augmentation problem as a generation-based graph OOD augmentation strategy, which directly models and augments the training distribution, without explicitly requiring the knowledge or separation of environmental information.
Specifically, we target on synthesizing an augmented training distribution $\tilde{P}_\mathrm{tr}(G, Y)$, which is combined with the original training distribution to obtain the classifier, stated as:
\begin{equation}
    \underset{f}{\min} \  \mathbb{E}_{(G,Y) \sim \{P_\mathrm{tr}(G, Y) \cup \tilde{P}_\mathrm{tr}(G, Y)\}}[\ell(f(G), Y)].
\end{equation}

The augmented distribution $\tilde{P}_\mathrm{tr}(G, Y)$ can be implemented in multiple ways, but it needs to satisfy two principles: 
(i) $\tilde{P}_\mathrm{tr}(G, Y)$ should deviate from $P_\mathrm{tr}(G, Y)$ in a controlled manner for exploration, and (ii) the explored graphs in $\tilde{P}_\mathrm{tr}(G, Y)$ should preserve the stable patterns of graphs in $P_\mathrm{tr}(G, Y)$. However, current graph generation models~\cite{jo2022score, martinkus2022spectre, vignac2022digress} cannot directly generate graphs that meet these two criteria. To address this, we propose a novel score-based conditional generative model that captures the augmented distribution $\tilde{P}_\mathrm{tr}(G, Y)$ while adhering to both principles.

\noindent\textbf{Motivation.} From the generative perspective, the goal of exploring the training distribution $P_{\mathrm{tr}}(G, Y)$ is to generate samples outside in-distribution from the conditional distribution $P_{\mathrm{tr}} \left(G, Y \mid\mathcal{E}_{\text {ood }}\right)$, where the exploration variable $\mathcal{E}_{\text {ood }}$ controls the OOD-ness of the generative process. 
The augmented distribution $\tilde{P}_{\mathrm{tr}}(G, Y)$ is then modelled by the conditional graph distribution $P_{\mathrm{tr}} \left(G, Y \mid\mathcal{E}_{\text {ood }}=\lambda\right)$, which can be decomposed as follows:
\begin{equation}
\label{eq:probability_aug}
 P_{\mathrm{tr}} \left(G, Y \mid\mathcal{E}_{\text {ood }}=\lambda\right)  \propto p\left(G\right) p\left(Y \mid G\right) p\left(\mathcal{E}_{\text {ood }}=\lambda \mid G, Y\right).
\end{equation}
Existing graph generation models~\cite{jo2022score, martinkus2022spectre, vignac2022digress, lee2023exploring} cannot directly sample graphs from the conditional distribution in Equation~(\ref{eq:probability_aug}), as it is infeasible to enumerate all possible $\mathcal{E}_{\text {ood }}$ values and their corresponding graphs and labels to compute the normalized probabilities. To overcome this limitation, we propose a novel score-based generative model to generate graphs from the target distribution.

\subsection{Preliminaries}
\label{secsec:prior_knowledge}

\textbf{Causal View.}
Following \cite{wu2022discovering}, we assume that each input graph $G$ can be decomposed into a causally relevant component $C$ and a spurious (environment-specific) component $S$, e.g., the House motif vs. the Tree base in GOOD-Motif~\cite{gui2022good}. The label $Y$ depends only on $C$, while $S$ captures variations across environments. These qualitative causal assumptions~\cite{pearl2016causal,neuberg2003causality} can be summarized as:
\begin{itemize}[leftmargin=*]
  \item $C \to G \leftarrow S$ \quad ($G$ consists of $C$ and $S$).
  \item $C \to Y$ \quad ($C$ is sufficient to determine $Y$).
  \item $C \not\!\perp\!\!\!\perp S$ \quad ($C$ and $S$ may be statistically dependent, e.g., via latent confounding).
  \item $S \perp\!\!\!\perp Y \mid C$ \quad (given $C$, $S$ carries no more information about $Y$).
\end{itemize}

\noindent\textbf{Low-Density Regions.}
Class-conditional densities are typically long-tailed \cite{sehwag2022generating}, with low-density neighborhoods on the data manifold containing few or no training samples. OOD graphs $G_{\text{ood}}$ tend to reside in these regions \cite{lee2023exploring}, and diffusion models are observed to interpolate rather than memorize in such regions~\cite{sehwag2022generating}. In \proposedmodel{}, we introduce an exploration parameter $\lambda$ to control the extent to which generation is encouraged toward these sparse regions.

Let $\tilde{P}_e$ denote the generator’s joint distribution at exploration level $e \equiv \mathcal{E}_{\text{ood}}(\lambda)$. When exploration is disabled ($e=0$), the generator should reproduce the in-distribution data:
\[
\tilde{P}_0(G,Y) \approx P_{\text{in}}(G,Y).
\]
As the exploration level increases, the generator should allocate more probability mass to OOD neighborhoods of the data manifold. Writing $\mathcal{G}_{\text{ood}}$ for a measurable subset of low-density OOD regions and taking $e_2>e_1\ge0$, we expect
\[
\tilde{P}_{e_2}\!\big(G \in \mathcal{G}_{\text{ood}}, Y\big)
\;>\;
\tilde{P}_{e_1}\!\big(G \in \mathcal{G}_{\text{ood}}, Y\big),
\]
that is, in shorthand,
\[
\tilde{P}(G_{\text{ood}},Y \mid \mathcal{E}_{\text{ood}}>0)
\;>\;
\tilde{P}(G_{\text{ood}},Y \mid \mathcal{E}_{\text{ood}}=0).
\]



\subsection{Forward Diffusion Process}
\begin{figure*}
  \centering
  \includegraphics[width=\textwidth]{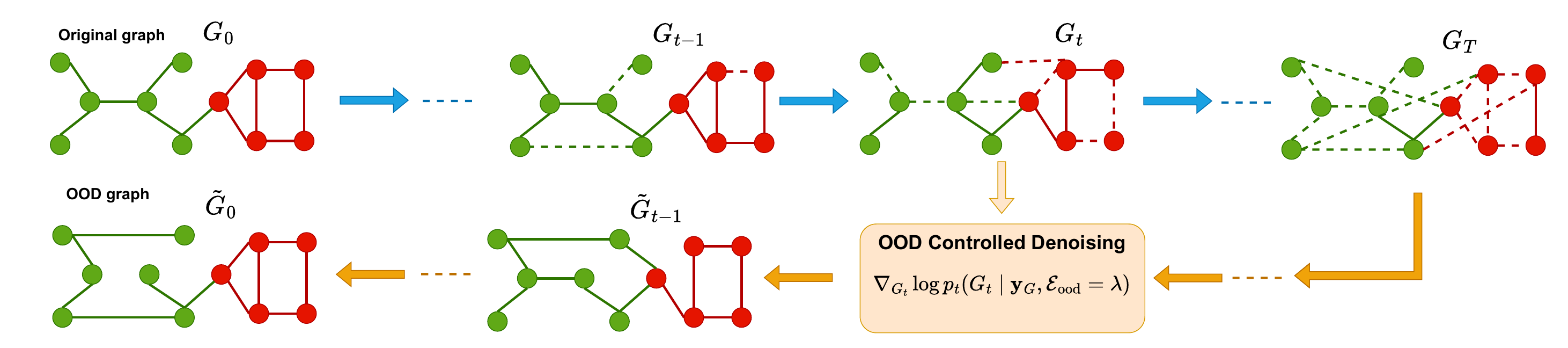}
  \caption{The diffusion and reverse processes of OODA. The diffusion process iteratively transforms an unlabeled original graph $G_0$ into noise $G_T$. During the denoising process, $\tilde{G}_{t-1}$ is computed using the conditional score $\nabla_{G_t} \log p_t\left(G_t \mid \mathbf{y}_G,\mathcal{E}_{\text {ood }}=\lambda\right)$, constrained by the target class $\mathbf{y}_G$ and the exploration parameter $\lambda$. Ultimately, the clean OOD graph $\tilde{G}_0$ is generated. 
  }
  \label{fig:ooda_schematic}
\end{figure*}

In this work, we focus on diffusion over continuous features to address distribution shifts that occur not only in discrete graph structures but also in continuous node and edge features. The foundational work~\cite{song2020score} introduced a method for modeling the diffusion process of data into noise and vice versa using stochastic differential equations (SDEs). For graph generation, this diffusion process gradually corrupts graphs into a prior distribution like the normal distribution. As illustrated in Figure~\ref{fig:ooda_schematic}, given an unlabeled graph $G$, we use continuous time $t \in[0, T]$ to index the diffusion steps $\left\{G_{t}\right\}_{t=1}^T$ of the graph, where $G_{0}$ represents the original distribution and $G_{T}$ follows a prior distribution. The forward diffusion process from the graph to the prior distribution is defined via an Itô SDE:
\begin{equation}
\label{eq:original_reverse_sde}
\mathrm{d} G_t=\mathbf{f}_t\left(G_t\right) \mathrm{d} t+g_t \mathrm{~d} \mathbf{w},
\end{equation}
which incorporates linear drift coefficient $\mathbf{f}_t(\cdot): \mathcal{G} \rightarrow \mathcal{G}$ and scalar diffusion coefficient $g_t: \mathcal{G} \rightarrow \mathbb{R}$ related to the amount of noise corrupting the unlabelled graph at
each infinitesimal step $t$, along with a standard Wiener process $\mathbf{w}$.

\subsection{Score-based Conditional Graph Generation with OOD Control}
\label{sec:score_ood_control}

To steer generation toward controlled OOD characteristics, we condition the reverse-time SDE on the target class \(\mathbf{y}_G\) and an exploration knob \(\lambda\in[0,1]\) that modulates OOD-ness:
\begin{equation}
\mathrm{d}G_t=\Big[\mathbf{f}_t(G_t)-g_t^2\,\nabla_{G_t}\log p_t\!\big(G_t,\mathbf{y}_G\mid \mathcal{E}_{\mathrm{ood}}=\lambda\big)\Big]\mathrm{d}\bar{t}
\;+\; g_t\,\mathrm{d}\overline{\mathbf{w}}.
\end{equation}
Since \(\mathbf{y}_G\) and \(\mathcal{E}_{\mathrm{ood}}\) are constants w.r.t.~\(G_t\) (detailed proof in Appendix~\ref{sec:proofs_sde}),
\begin{equation}
\nabla_{G_t}\log p_t\!\big(G_t,\mathbf{y}_G\mid \mathcal{E}_{\mathrm{ood}}=\lambda\big)
=\nabla_{G_t}\log p_t\!\big(G_t\mid \mathbf{y}_G,\mathcal{E}_{\mathrm{ood}}=\lambda\big).
\end{equation}
Thus the conditional reverse-time SDE becomes
\begin{equation}
\label{eq:conditional_sde}
\mathrm{d}G_t=\Big[\mathbf{f}_t(G_t)-g_t^2\,\nabla_{G_t}\log p_t\!\big(G_t\mid \mathbf{y}_G,\mathcal{E}_{\mathrm{ood}}=\lambda\big)\Big]\mathrm{d}\bar{t}
\;+\; g_t\,\mathrm{d}\overline{\mathbf{w}}.
\end{equation}

\noindent\textbf{Score Decomposition.}
By Bayes’ rule,
\begin{equation}
\label{eq:sum_score_componets}
\begin{aligned}
& \nabla_{G_t} \log p_t\left(G_t \mid \mathbf{y}_G,\mathcal{E}_{\text {ood }} =\lambda\right) \\ & = \nabla_{G_t} \log p_t\left(G_t\right) + \nabla_{G_t} \log p_t\left(\mathbf{y}_G \mid G_t\right) \\ 
& +\nabla_{G_t} \log p_t\left(\mathcal{E}_{\text {ood }}=\lambda \mid G_t, \mathbf{y}_G\right).
\end{aligned}
\end{equation}
where the last term, the \emph{OOD-ness score}, is intractable to model directly. This is because modeling the conditional distribution, $$p_t\left(\mathcal{E}_{\text {ood }}=\lambda \mid G_t, \mathbf{y}_G\right)$$ would require an infeasible enumeration over a continuous OOD space and all corresponding graphs and labels.

\noindent\textbf{A Tractable OOD Surrogate.}
From the causal view in \S\ref{secsec:prior_knowledge}, OOD graphs reside in low-density regions of the training joint \(P_{\mathrm{tr}}(G,Y)\). We encode this by assigning higher OOD probability to lower joint density and posit the surrogate
\begin{align}
p_t\left(\mathcal{E}_{\text {ood }}=\lambda \mid G_t, \mathbf{y}_G\right) & \propto  p_t\left(G_t, \mathbf{y}_G\right)^{-\sqrt{\lambda}}\\
&=\big[p_t(G_t)\,p_t(\mathbf{y}_G\mid G_t)\big]^{-\sqrt{\lambda}}.\nonumber
\end{align}
which yields
\begin{equation}
\begin{aligned}
& \nabla_{G_t}\log p_t\!\big(\mathcal{E}_{\mathrm{ood}}=\lambda \mid G_t,\mathbf{y}_G\big)
= \\
& -\sqrt{\lambda}\,\Big(\nabla_{G_t}\log p_t(G_t)
+ \nabla_{G_t}\log p_t(\mathbf{y}_G\mid G_t)\Big).
\end{aligned}
\end{equation}
Plugging into \eqref{eq:sum_score_componets} gives a simple and tractable  conditional score:
\begin{align}
\label{eq:eq_9}
& \nabla_{G_t} \log p_t\left(G_t \mid \mathbf{y}_G,\mathcal{E}_{\text {ood }}=\lambda\right) = \\
& (1-\sqrt{\lambda})\Big(\nabla_{G_t}\log p_t(G_t)
+ \nabla_{G_t}\log p_t(\mathbf{y}_G\mid G_t)\Big).\nonumber 
\end{align}
We use \(0\le\lambda\le 1\) so that the guidance stays nonnegative and smoothly trades off between in-distribution sampling (\(\lambda=0\)) and exploration toward low-density regions (\(\lambda\to 1\)).

\noindent\textbf{Learning Components.} The tractable conditional score in Equation~(\ref{eq:eq_9}) necessitates estimating two key components: the unconditional score $\nabla_{G_t} \log p_t\left(G_t\right)$ and the class-conditional probability $p_t\left(\mathbf{y}_G\mid G_t\right)$. To this end, we first train a score network $\boldsymbol{s}_{\theta, t}\left(G_t\right)$ which takes the noisy graph $G_t$ and time $t$ as inputs to approximate the unconditional score $\nabla_{G_t} \log p_t\left(G_t\right)$. Concurrently, we train a time-dependent classifier $\phi_t(G_t)$ to approximate the conditional class distribution $p_t\left(\mathbf{y}_G\mid G_t\right)$ using inputs $G_t$ and $t$. The classifier is trained on \emph{noisy} pairs \((G_t,\mathbf{y}_G)\) to capture time-dependent feature dynamics, rather than on clean \((G_0,\mathbf{y}_G)\).

\subsection{Working Principles of \proposedmodel}
\label{sec:working_principles}

Substituting \eqref{eq:eq_9} into \eqref{eq:conditional_sde} defines the full sampler. The single hyperparameter \(\lambda\) controls the amount of OOD exploration while retaining the class semantics via \(\mathbf{y}_G\).

\noindent\textbf{Assumptions (from \S\ref{secsec:prior_knowledge}).}
Each graph \(G\) decomposes into \(C\) (causal) and \(S\) (non-causal/spurious), with
\(C\to Y\), \(C\not\!\perp\!\!\!\perp S\), and \(S \perp\!\!\!\perp Y \mid C\).
In particular, \(p_t(\mathbf{y}_G\mid G_t)=p_t(\mathbf{y}_G\mid C_t)\).

\begin{proposition}[Causal structure preservation]
\label{prop:preserve_causal}
For $0\le \lambda\le 1$, the generation induced by
\eqref{eq:conditional_sde}–\eqref{eq:eq_9} preserves the causal
relation $C\!\to\!Y$ under the generated joint $\tilde{P}_\lambda(G,Y)$; equivalently,
$\tilde{P}_\lambda(Y\mid G)=\tilde{P}_\lambda(Y\mid C)$ for all $\lambda\in[0,1]$.
\end{proposition}

\noindent\textbf{Proof.}
\textbf{Assumptions (cf.\ Section~\ref{secsec:prior_knowledge}).}
Each graph decomposes as $G=(C,S)$ with $S\perp\!\!\!\perp Y\mid C$ and hence
$P(Y\mid G)=P(Y\mid C)$. The time-dependent classifier is trained on noisy pairs so that
$p_t(\mathbf{y}_G\mid G_t)=p_t(\mathbf{y}_G\mid C_t)$, implying
$\nabla_{S_t}\log p_t(\mathbf{y}_G\mid G_t)=0$.

\noindent\textbf{(i) Vector-Field Structure.}
From \eqref{eq:eq_9},
\begin{equation}
\begin{aligned}
& \nabla_{G_t}\log p_t\!\big(G_t\mid \mathbf{y}_G,\mathcal{E}_{\mathrm{ood}}=\lambda\big)
= \\
& (1-\sqrt{\lambda})\Big(\underbrace{\nabla_{G_t}\log p_t(G_t)}_{\text{label-free}}
+\underbrace{\nabla_{G_t}\log p_t(\mathbf{y}_G\mid G_t)}_{\text{$C$-only}}\Big).
\end{aligned}
\end{equation}
Projecting onto $S$-coordinates gives
$\nabla_{S_t}\log p_t(\mathbf{y}_G\mid G_t)=0$, so the label-guidance field has
no component on $S$. Thus the $S$-dynamics are label-independent for all $\lambda$.

\noindent\textbf{(ii) $\lambda=0$ anchors $C\!\to\!Y$.}
At $\lambda=0$, \eqref{eq:eq_9} reduces to standard class-conditional
guidance $\nabla\log p_t(G_t)+\nabla\log p_t(\mathbf{y}_G\mid G_t)$ targeting
$P_{\mathrm{tr}}(G\mid Y)$; consequently
$\tilde{P}_0(\tilde{C}\mid \mathbf{y}_G=y)\approx P_{\mathrm{tr}}(C\mid y)$ and
$\tilde{P}_0(Y\mid G)=\tilde{P}_0(Y\mid C)$.

\noindent\textbf{(iii) Continuation in $\lambda$.}
For $0<\lambda\le 1$, the overall field is scaled by $(1-\sqrt{\lambda})\ge 0$ and
the label term remains $C$-only. Since the $S$-evolution is label-free, changing
$\lambda$ alters how broadly we explore in $S$ but cannot introduce dependence of
$Y$ on $S$ given $C$. Hence $P(Y\mid G)=P(Y\mid C)$ is preserved along the
reverse-time flow for any $\lambda\in[0,1]$.

Therefore, for all $\lambda\in[0,1]$, $\tilde{P}_\lambda(Y\mid G)=\tilde{P}_\lambda(Y\mid C)$,
i.e., the causal relation $C\to Y$ is preserved.

\begin{proposition}[Spurious pattern reduction]
\label{prop:reduce_spurious}
Let $\tilde{P}_\lambda$ be the generated joint induced by \eqref{eq:eq_9}.
For any fixed class $y$, write
$\tilde{\pi}_\lambda(s\mid y):=\tilde{P}_\lambda(S=s\mid \mathbf{y}_G=y)$ and
$\pi_0(s\mid y):=P_{\mathrm{tr}}(S=s\mid Y=y)$.
Then for $0\le \lambda_1<\lambda_2\le 1$ and any measurable “top-mass” set
$\mathcal{S}_\tau(y):=\{\,s:\pi_0(s\mid y)\ge \tau\,\}$, it holds that
\[
\tilde{\pi}_{\lambda_2}\big(\mathcal{S}_\tau(y)\mid y\big)
\;\le\;
\tilde{\pi}_{\lambda_1}\big(\mathcal{S}_\tau(y)\mid y\big).
\]
In words: as $\lambda$ increases, the probability mass assigned to training-typical
spurious patterns $S$ monotonically decreases under $\tilde{P}_\lambda$.
\end{proposition}

\noindent\textbf{Proof.}
\textbf{(i) Semantics are $C$-anchored.}
By Section~\ref{secsec:prior_knowledge}, $S\perp\!\!\!\perp Y\mid C$, hence $P(Y\mid G)=P(Y\mid C)$.
Moreover the time-dependent classifier is trained on noisy pairs so that
$p_t(\mathbf{y}_G\mid G_t)=p_t(\mathbf{y}_G\mid C_t)$, implying
$\nabla_{S_t}\log p_t(\mathbf{y}_G\mid G_t)=0$.
Thus the label-guidance term in \eqref{eq:eq_9} has no component on $S_t$.

\noindent\textbf{(ii) Low-density reweighting induces power tempering.}
With the OOD surrogate
$p_t(\mathcal{E}_{\mathrm{ood}}=\lambda\mid G_t,\mathbf{y}_G)\propto p_t(G_t,\mathbf{y}_G)^{-\sqrt{\lambda}}$,
the class-conditional sampler satisfies
\[
p_t(G_t\mid \mathbf{y}_G,\mathcal{E}_{\mathrm{ood}}=\lambda)
\;\propto\; p_t(G_t\mid \mathbf{y}_G)^{\alpha(\lambda)},
\]
\[
\alpha(\lambda):=1-\sqrt{\lambda}\in(0,1].
\]
Marginalizing $C_t$ gives, for any fixed $y$,
\[
\tilde{\pi}_\lambda(s\mid y)\;\propto\;
\int p_t(C_t,s\mid y)^{\alpha(\lambda)}\,\mathrm{d}C_t,
\]
followed by normalization over $s$.
Note that $\alpha(\lambda)$ decreases as $\lambda$ increases.

\noindent\textbf{(iii) Power tempering compresses high-probability advantages.}
Let $q_s:=\int p_t(C_t,s\mid y)\,\mathrm{d}C_t=\pi_0(s\mid y)$.
For any $s_1,s_2$ with $q_{s_1}\ge q_{s_2}>0$ and any $\alpha\in(0,1]$,
\[
\frac{q_{s_1}^{\alpha}}{q_{s_2}^{\alpha}}
=\Big(\frac{q_{s_1}}{q_{s_2}}\Big)^{\alpha}
\le \frac{q_{s_1}}{q_{s_2}}.
\]
Hence, as $\alpha(\lambda)$ decreases (i.e., $\lambda$ increases), the normalized
mass shifts away from high-density (“typical”) $S$ configurations toward lower-density ones.

\noindent\textbf{(iv) Conclusion.}
Combining (i)–(iii), increasing $\lambda$ cannot alter the $C$-anchored semantics,
but it reduces the normalized probability of training-typical $S$ under
$\tilde{\pi}_\lambda(\cdot\mid y)$, yielding the stated monotonicity on any
top-mass set $\mathcal{S}_\tau(y)$.

\section{Experiments}
\label{sec:experiments}
We first demonstrate the effectiveness of our diffusion models on graph OOD tasks. Then, we validate that the desired augmentation principles are successfully realized. Additional experiments regarding sensitivity of hyperparameter $\lambda$, as well as time and memory complexity, are presented in Appendix~\ref{sec:sensitivity_lambda} and~\ref{sec:time_complex}.

\subsection{Experimental Settings}
\label{secsec:experimental settings}
\textbf{Setup. } We adopt the same evaluation metrics as in~\cite{gui2022good} for a fair comparison. The model that achieves the best performance on the OOD validation sets is then evaluated on the OOD test sets. Furthermore, to ensure fair comparison across all methods, we utilize the same GNN backbones—GIN~\cite{xu2018how} and GIN-Virtual~\cite{xu2018powerful, gilmer2017neural}—as applied in the GOOD benchmark~\cite{gui2022good} for each dataset. The experimental details, including evaluation metrics and hyperparameter configurations, are summarized in Appendix~\ref{sec:random_gin} and~\ref{sec:experimental_details}.

\begin{table*}

  \caption{Performance on synthetic and real-world datasets. Bold numbers indicate the best performance, while the underlined numbers indicate the second best performance. Bootstrap test is used to assess significance with $20000$ resamples. The dagger symbol $\left({ }^{\dagger}\right)$ indicates that our method significantly outperforms the best baseline with a p-value less than $0.05$.
  }
  \label{tab:main_results}
  \begin{center}
  \resizebox{\textwidth}{!}{
    \begin{tabular}{ccccccccc}
        \toprule
        \multirow{2}{*}{Type} & \multirow{2}{*}{Method} & \multicolumn{2}{c}{Motif} & \multicolumn{1}{c}{CMNIST} & \multicolumn{2}{c}{Molhiv} & \multicolumn{1}{c}{GOOD-SST2} \\ \cmidrule{3-9} 
         &  & \multicolumn{1}{c}{base} & \multicolumn{1}{c}{size} & \multicolumn{1}{c}{color} & scaffold & \multicolumn{1}{c}{size} & length \\ \midrule
        \multirow{6}{*}{General Generalization} & ERM & \multicolumn{1}{c}{$68.66 \pm 4.25$} & \multicolumn{1}{c}{$51.74\pm 2.88$} & \multicolumn{1}{c}{$28.60\pm 1.87$} & $69.58 \pm 2.51$ & $59.94 \pm 2.37$ & \multicolumn{1}{c}{$81.30 \pm 0.35$} &  \\
         & IRM & \multicolumn{1}{c}{$70.65\pm 4.17$} & \multicolumn{1}{c}{$51.41 \pm 3.78$} & \multicolumn{1}{c}{$27.83 \pm 2.13$} & $67.97 \pm 1.84$ & $59.00 \pm 2.92$ & \multicolumn{1}{c}{$79.91 \pm 1.97$} &  \\
         & GroupDRO & \multicolumn{1}{c}{$68.24 \pm 8.92$} & \multicolumn{1}{c}{$51.95 \pm 5.86$} & \multicolumn{1}{c}{$29.07 \pm 3.14$} & $70.64 \pm 2.57$ & $58.98 \pm 2.16$ & \multicolumn{1}{c}{$81.35 \pm 0.54$} &  \\
         & VREx & \multicolumn{1}{c}{$71.47 \pm 6.69$} & \multicolumn{1}{c}{$52.67 \pm 5.54$} & \multicolumn{1}{c}{$28.48 \pm 2.87$} & $70.77 \pm 2.84$ & $58.53 \pm 2.88$ & \multicolumn{1}{c}{$80.64 \pm 0.35$} &  \\
         & DANN & \multicolumn{1}{c}{$65.47 \pm 5.35$} & \multicolumn{1}{c}{$51.46 \pm 3.41$} & \multicolumn{1}{c}{$29.14 \pm 2.93$} & $70.63 \pm 1.82$ & $\underline{62.38 \pm 2.65}$ & \multicolumn{1}{c}{$79.71 \pm 1.35$} &  \\ 
         & Deep Coral & \multicolumn{1}{c}{$68.88 \pm 3.61$} & \multicolumn{1}{c}{$53.71 \pm 2.75$} & \multicolumn{1}{c}{$29.05 \pm 2.19$} & $68.61 \pm 1.70$ & $60.11 \pm 3.53$ & \multicolumn{1}{c}{$79.81 \pm 0.22$} &  \\\midrule
        \multirow{3}{*}{Graph Generalization} & DIR & \multicolumn{1}{c}{$62.07 \pm 8.75$} & \multicolumn{1}{c}{$52.27 \pm 4.56$} & \multicolumn{1}{c}{$33.20 \pm 6.17$} & $68.07 \pm 2.29$ & $58.08 \pm 2.31$ & \multicolumn{1}{c}{$77.65 \pm 1.93$} &  \\
         & GSAT & $62.80 \pm 11.41$ & $53.20 \pm 8.35$  & $28.17 \pm 1.26$ & $68.66 \pm 1.35$ & $58.06 \pm 1.98$ & $81.49 \pm 0.76$ &  \\
         & CIGA & $66.43 \pm 11.31$ & $49.14\pm 8.34$ & $32.22 \pm 2.67$ & $69.40 \pm 2.39$ & $59.55 \pm 2.56$ & $80.44 \pm 1.24$ &  \\\midrule
        \multicolumn{1}{l}{\multirow{8}{*}{Graph Augmentation}} & DropNode & $\underline{74.55 \pm 5.56}$ & $54.14 \pm 3.11$ & $33.01 \pm 0.12$ & $\underline{71.18 \pm 1.16}$ & $58.52 \pm 0.49$ & $81.14 \pm 1.73$ &  \\
        & DropEdge & $45.08 \pm 4.46$ & $45.63 \pm 4.61$ & $22.65 \pm 2.90$ & $70.78 \pm 1.38$ & $58.53 \pm 1.26$ & $78.93 \pm 1.34$ &  \\
        & MaskFeature & $64.98 \pm 6.95$ & $52.24 \pm 3.75$ & $44.85 \pm 2.42$ & $65.90 \pm 3.68$ & $62.30 \pm 3.17$ & $\underline{82.00 \pm 0.73}$ &  \\
        \multicolumn{1}{l}{} & FLAG & $61.12 \pm 5.39$ & $51.66 \pm 4.14$ & $32.30 \pm 2.69$ & $68.45 \pm 2.30$ & $60.59 \pm 2.95$ & $77.05 \pm 1.27$ &  \\
        \multicolumn{1}{l}{} & M-Mixup & $70.08 \pm 3.82$ & $51.48 \pm 4.91$ & $26.47 \pm 3.45$ & $68.88 \pm 2.63$ & $59.03 \pm 3.11$ & $80.88 \pm 0.60$ &  \\
        \multicolumn{1}{l}{} & G-Mixup & $59.66 \pm 7.03$ & $52.81 \pm 6.73$ & $31.85 \pm 5.82$ & $70.01 \pm 2.52$ & $59.34 \pm 2.43$ & $80.28 \pm 1.49$ &  \\
        \multicolumn{1}{l}{} & GRATIN & $72.45 \pm 4.23$ & $52.37 \pm 3.17$ & $32.77 \pm 0.50$ & $70.23 \pm 2.29$ & $60.28 \pm 2.12$ & $80.64 \pm 0.51$ &  \\
        \multicolumn{1}{l}{} & AIA & $73.64 \pm 5.15$ & $\underline{55.85 \pm 7.98}$ & $\underline{36.37 \pm 4.44}$ & $71.15 \pm 1.81$ & $61.64 \pm 3.37$ & $81.69 \pm 0.57$ &  \\
        \multicolumn{1}{l}{} & \textbf{\proposedmodel{}(Ours)} & $\mathbf{75.25 \pm 3.84}^{\dagger}$  & $\mathbf{60.81 \pm 7.80}^{\dagger}$ & $\mathbf{54.60 \pm 2.27}^{\dagger}$ & $\mathbf{72.67 \pm 1.28}^{\dagger}$  & $\mathbf{63.66 \pm 1.20}^{\dagger}$ & $\mathbf{82.69 \pm 0.28}^{\dagger}$ &  \\ \bottomrule
    \end{tabular}
    }
    \end{center}

\end{table*}

\noindent\textbf{Datasets.}
We evaluate on synthetic, semi-artificial, and real-world benchmarks from GOOD~\cite{gui2022good}: \textsc{GOOD-Motif}, \textsc{GOOD-CMNIST}, \textsc{GOOD-HIV}, and \textsc{GOOD-SST2}. Following the official protocol, we use the \emph{base}, \emph{size}, \emph{color}, \emph{scaffold}, and \emph{length} splits to induce diverse covariate shifts in structure, node features, and edge features. Although some datasets are not Web-native, the shifts emulate common Web scenarios: motif and size shifts mirror evolving community wiring and graph growth in social/hyperlink networks; color shifts stand in for feature/style changes (e.g., UI or content modality drift); scaffold splits mimic cold-start categories/items in e-commerce or knowledge graphs; and length shifts capture post/review length drift in social and content platforms. These settings therefore serve as Web-analogous stress tests for OOD generalization on graphs. Full dataset statistics, preprocessing details, and split constructions are provided in the Appendix~\ref{sec:experimental_details}.

\noindent\textbf{Baselines.} We adopt 17 baselines, which can be divided into the following three specific categories:(i) \emph{general generalization algorithms}, including ERM, IRM~\cite{arjovsky2019invariant}, GroupDRO~\cite{sagawa2019distributionally}, VREx~\cite{krueger2021out}, DANN~\cite{ganin2016domain}, Deep Coral~\cite{sun2016deep}; (ii) \emph{graph generalization algorithms}, including DIR~\cite{wu2022discovering}, GSAT~\cite{miao2022interpretable}, CIGA~\cite{chen2022learning}, and (iii) \emph{graph data augmentation techniques}, including DropNode~\cite{feng2020graph}, DropEdge~\cite{rong2019dropedge}, MaskFeature~\cite{thakoor2021large}, FLAG~\cite{kong2022robust}, M-Mixup~\cite{wang2021mixup}, G-Mixup~\cite{han2022g}, AIA~\cite{sui2024unleashing}, GRATIN~\cite{abbahaddou2025graph}.

\subsection{Graph Out-of-Distribution Classification}
\label{sec:ood_classification_results}
The graph classification performances under covariate shift are presented in Table~\ref{tab:main_results}. As shown, \proposedmodel{} consistently outperforms all baseline methods across diverse covariate shifts and different datasets. On the synthetic dataset GOOD-Motif, \proposedmodel{} achieves a performance improvement of $6.59\%$ over ERM under base shift and $9.07\%$ under size shift. For the semi-artificial dataset GOOD-CMNIST, designed for node feature shifts, performance is significantly enhanced by $18.23\%$ compared to the leading graph augmentation method, AIA, and improved by $21.40\%$ over the best graph invariant learning method, DIR. In the real-world molecular dataset GOOD-HIV, where covariate shifts occur in graph structure, node features, and edge features simultaneously, \proposedmodel{} outperforms ERM by $3.09\%$ on scaffold shift and by $3.72\%$ on size shift. For the real-world sentiment analysis dataset GOOD-SST2, while AIA is outperformed by MaskFeature by $0.31\%$, \proposedmodel{} exceeds MaskFeature by $0.69\%$.

We assess statistical significance using a paired nonparametric bootstrap test~\cite{davison1997bootstrap} at the $5\%$ level to compare \proposedmodel{} with leading baselines; details are in Appendix~\ref{sec:t_test}. Overall, these results demonstrate that no baseline graph invariant learning or graph data augmentation methods consistently outperform each other under various covariate shifts. \proposedmodel{} enhances environmental exploration by generating OOD graphs in a controlled manner while preserving stable features. Consequently, \proposedmodel{} reliably improves performance across different datasets facing various covariate shifts.

\subsection{OOD Controlled Graph Generation}
\label{sec:ood_generation_results}
In this section, we present both qualitative and quantitative experiments to demonstrate the effectiveness of our OOD diffusion augmentation framework. The experiments are conducted using the GOOD-Motif-base, GOOD-SST2-length and GOOD-HIV-scaffold datasets. Additionally, we present a comparative analysis of \proposedmodel{} against various augmentation methods, evaluating their respective capacities to explore environments while preserving label-determining stable structures. We further provides visualizations of the OOD graphs generated by \proposedmodel{} for the GOOD-Motif-basis and GOOD-Molhiv-scaffold datasets.


\noindent\textbf{Training Distribution Augmentation with Control.}
We first validate that our framework can explore the space of the original distribution $P_{\mathrm{tr}}(G, Y)$ and generate an augmented distribution $\tilde{P}_{\mathrm{tr}}(G, Y)$ in a controlled manner. To quantify deviation from training graphs, we adopt the GIN-based metrics of~\cite{thompson2022on}, which are expressive and computationally efficient for graph generators. Following their recommendation, we report Maximum Mean Discrepancy (MMD) with an RBF kernel as a robust measure of distributional shift and diversity. We vary $\lambda \in [0,1)$ in steps of $0.1$ to generate ten augmented sets (each the same size as the training set) and compute MMD (RBF) between each augmented set and the original training set. Estimation details are in the Appendix.

\textit{GOOD-Motif-basis.}
Figure~\ref{fig:covariate_shift_comparisons}(a) shows that MMD (RBF) between $P_{\mathrm{tr}}(G,Y)$ and $\tilde{P}_{\mathrm{tr}}(G,Y)$ increases monotonically with $\lambda$. At $\lambda=0.0$, MMD is $0.072 \pm 0.002$; by $\lambda=0.9$, it is approximately $1.5\times$ larger. Thus, \proposedmodel{} generates OOD graphs in a tunable manner by adjusting $\lambda$, while also increasing diversity as the exploration strength grows.

\textit{GOOD-SST2-length.}
Here, $\lambda$ controls OOD \emph{size} (e.g., $\lambda=0.1$ adds one unit to graph size relative to training). We again sweep $\lambda$ in $[0,1)$ by $0.1$ to obtain ten augmented sets. As shown in Figure~\ref{fig:covariate_shift_comparisons}(b), MMD (RBF) rises with $\lambda$: from $0.032 \pm 0.005$ at $\lambda=0.0$ to roughly $25\times$ that value at $\lambda=0.9$. 

\noindent\textbf{Stable Pattern Preservation.}
We use a pretrained graph transformer \(\phi_t\), trained on noisy graphs from \(P_{\mathrm{tr}}\), as a proxy to assess whether augmented graphs retain label-determining stable patterns. Specifically, we compute \(p(\mathbf{y}_G \mid \tilde{G}_0)\) and report \(\mathbb{E}_{\tilde{G}_0 \sim \tilde{P}}[p(\mathbf{y}_G \mid \tilde{G}_0)]\). On GOOD-Motif-basis (Figure~\ref{fig:covariate_shift_comparisons}(c)), this expectation remains above \(0.95\) across \(\lambda\); on GOOD-SST2-length (Figure~\ref{fig:covariate_shift_comparisons}(d)), it exceeds \(0.94\). These results indicate that \proposedmodel{} generates OOD graphs while preserving stable, label-determining structure.

\noindent\textbf{Comparison with augmentation baselines.}
Figure~\ref{fig:covariate_shift_comparisons} compares methods on (i) environmental exploration and (ii) preservation of stable structures. G-Mixup mixes continuous labels, making discrete structure preservation ill-posed; we therefore omit that metric for G-Mixup. Chemistry-specific metrics (e.g., FCD~\cite{preuer2018frechet}, RDKit-based scores~\cite{landrum2016rdkit}) require valid discrete molecules and are incompatible with baselines that output continuous graph features (e.g., AIA), so we focus on general graph metrics. Across GOOD-Motif-basis and GOOD-SST2, DropEdge explores more aggressively than G-Mixup and AIA but often distorts stable patterns. In contrast, \proposedmodel{} achieves stronger, \emph{tunable} exploration via \(\lambda\) (Figure~\ref{fig:covariate_shift_comparisons}(a) and Figure~\ref{fig:covariate_shift_comparisons}(b)) while consistently preserving stable structures better than the baselines (Figure~\ref{fig:covariate_shift_comparisons}(c) and Figure~\ref{fig:covariate_shift_comparisons}(d)).

\begin{figure}[t]
  \centering

  \begin{minipage}[t]{0.48\columnwidth}
    \centering
    \includegraphics[width=\linewidth]{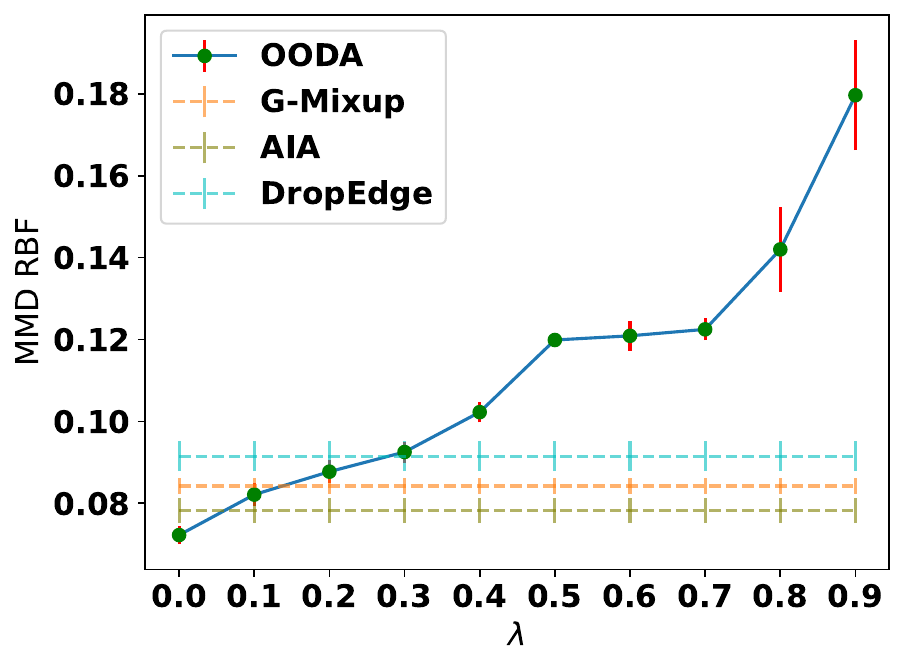}\\[-0.8em]
    \vspace{2pt}
    \panelxy{1.5em}{-0.0ex}{a}
  \end{minipage}\hfill
    \begin{minipage}[t]{0.48\columnwidth}
    \centering
    \includegraphics[width=\linewidth]{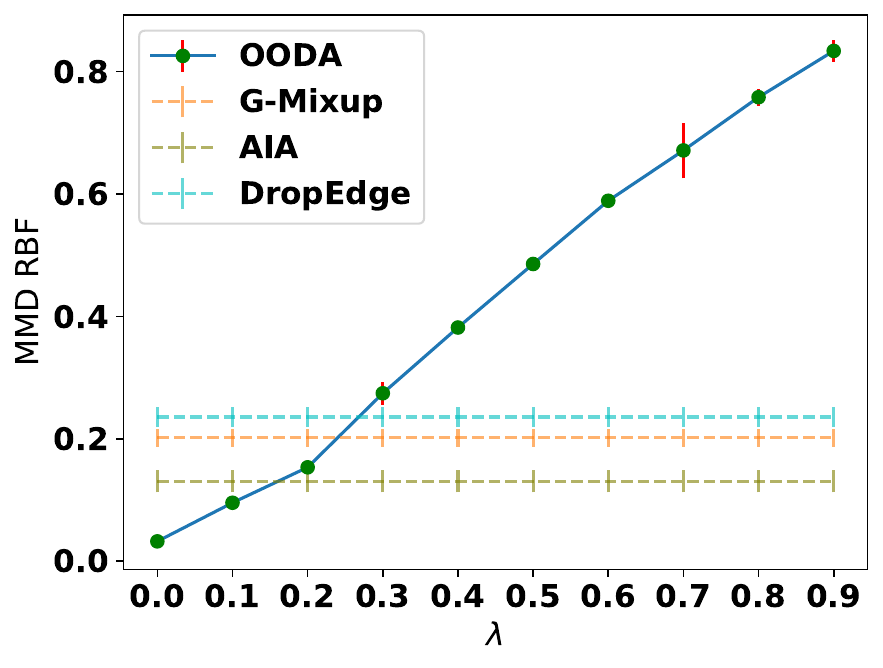}\\[-0.8em]
    \vspace{2pt}
    \panelxy{1.5em}{-0.2ex}{b}
  \end{minipage}


    \begin{minipage}[t]{0.48\columnwidth}
    \centering
    \includegraphics[width=\linewidth]{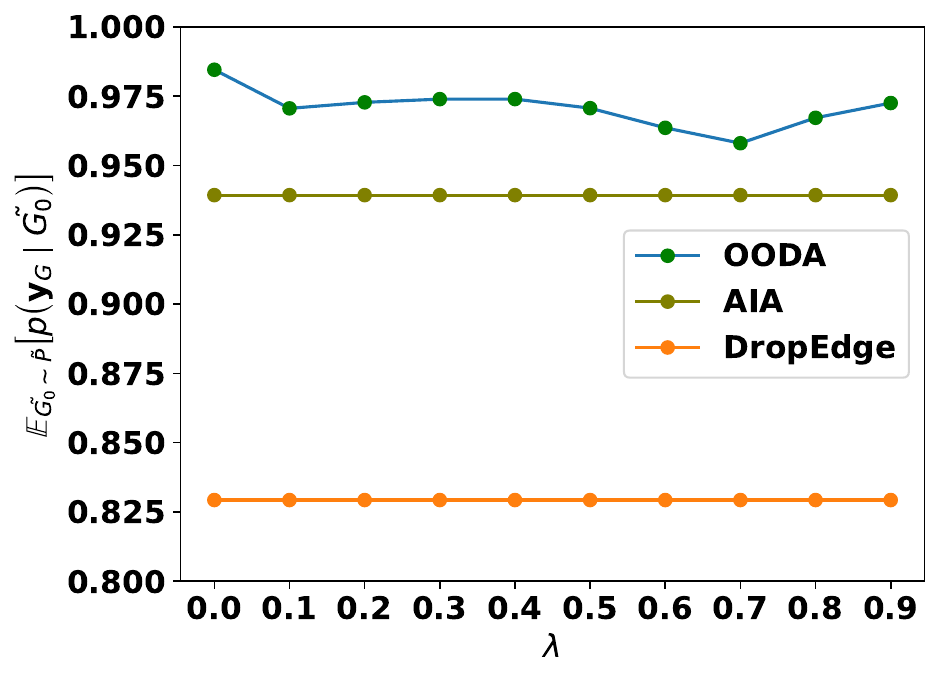}\\[-0.8em]
    \vspace{2pt}
    \panelxy{1.5em}{-0.2ex}{c}
  \end{minipage} \hfill
  \begin{minipage}[t]{0.48\columnwidth}
    \centering
    \includegraphics[width=\linewidth]{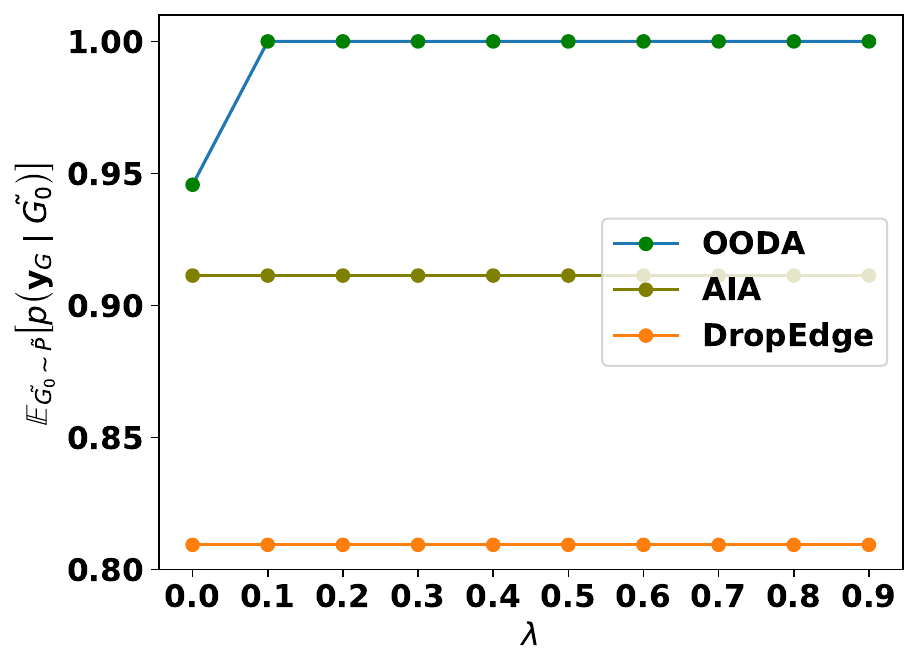}\\[-0.8em]
    \vspace{2pt}
    \panelxy{1.5em}{-0.2ex}{d}
  \end{minipage}

  \caption{
    Augmentation comparison on GOOD-Motif-basis and GOOD-SST2-length. 
    (a) GOOD-Motif-basis: distribution distance between $P_{\mathrm{tr}}(G,Y)$ and $\tilde P_{\mathrm{tr}}(G,Y)$ (MMD-RBF).
    (b) GOOD-SST2-length: distribution distance (MMD-RBF). 
    (c) GOOD-Motif-basis: stable pattern preservation probability (probability augmented graphs preserve label-determining stable patterns). 
    (d) GOOD-SST2-length: stable pattern preservation probability.
  }
  \label{fig:covariate_shift_comparisons}
\end{figure}

\begin{figure}[ht]  
    \begin{center}
    \begin{subfigure} 
        \centering
        \includegraphics[scale=0.25]{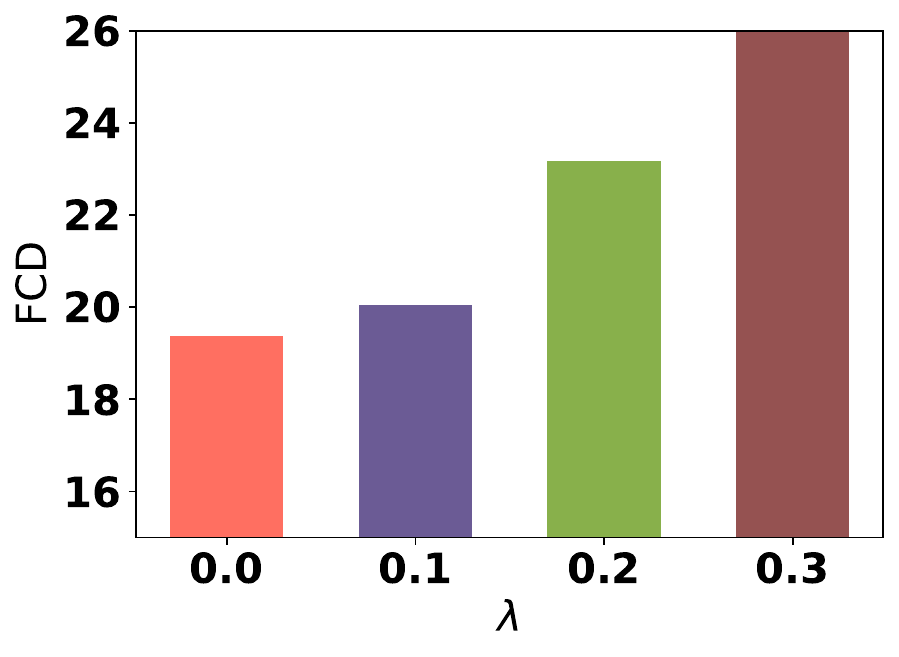}
    \end{subfigure}
    \hspace{0.2cm}
    \begin{subfigure} 
        \centering
        \includegraphics[scale=0.25]{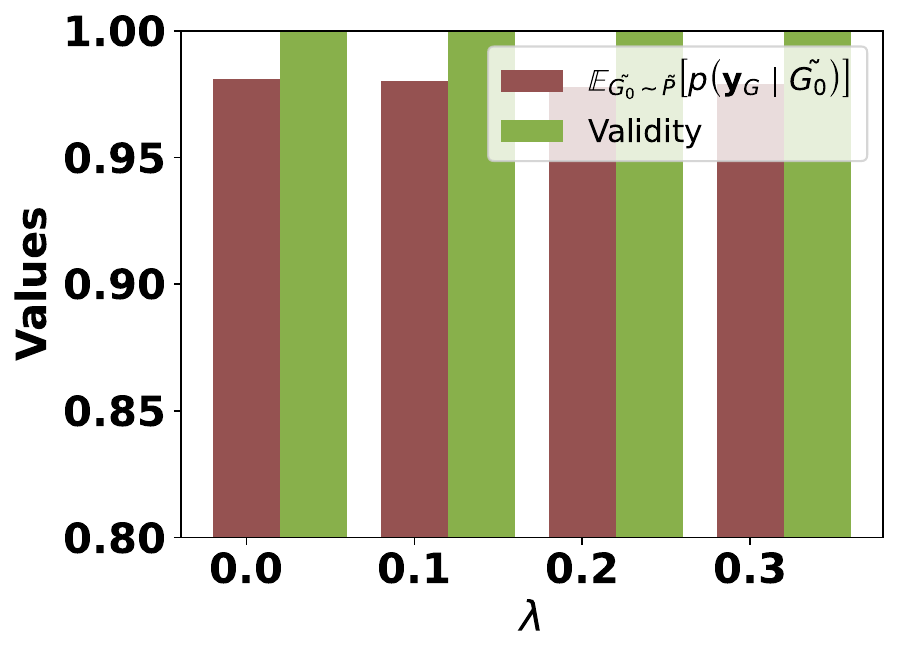}
    \end{subfigure}
    \caption{(Left): Distance between the original GOOD-HIV graph distribution and the augmented graph distribution. (Right): The validity of the OOD molecules and the expected probabilities that the OOD GOOD-HIV molecules retain stable patterns.}
    \label{fig:mol_fcd_cons}
    \end{center}
\end{figure}

\noindent\textbf{Validity of OOD Molecule Generation. }
To further verify the effectiveness of \proposedmodel{} in generating valid OOD molecules, we assess both the exploration of new patterns and the preservation of stable features. The Fréchet ChemNet Distance (FCD)~\cite{preuer2018frechet} is employed to quantify the distance between the training and augmented distributions of molecules, based on the penultimate activations of ChemNet. Additionally, RDKit~\cite{landrum2016rdkit} is used to evaluate the fraction of valid molecules. We also compute the $\mathbb{E}_{\tilde{G_0} \sim \tilde{P}}\left[p\left(\mathbf{y}_G \mid \tilde{G_0}\right)\right]$ to confirm that the OOD molecules retain stable patterns necessary for inhibiting HIV replication. The results, shown in Figure~\ref{fig:mol_fcd_cons}, demonstrate that as $\lambda$ increases, the FCD between the training molecules and the generated OOD molecules grows. Despite this, the OOD molecules consistently preserve stable patterns, with $\mathbb{E}_{\tilde{G_0} \sim \tilde{P}}\left[p\left(\mathbf{y}_G \mid \tilde{G_0}\right)\right]$ remaining above $0.97$, while maintaining a $100\%$ validity rate.

\noindent\textbf{Ablation study. }
We present experiments to evaluate the impact of exploration guidance ($\lambda$) and stable patterns preservation guidance ($\alpha$). The results are summarized in Table~\ref{tab:ablation_results}. As shown, incorporating $\alpha$ guidance improves the OOD performance of the diffusion model trained on unlabeled graphs by ensuring that the generated graphs retain the stable patterns that determine the labels. Without $\alpha$ guidance, a diffusion model guided only by $\lambda$ tends to push the augmented graphs into arbitrary OOD regions, which negatively impacts performance to some extent. Ultimately, the combination of both $\alpha$ and $\lambda$ guidance enables the augmented distribution to capture both stable patterns and novel environmental patterns, resulting in the best overall performance.

\begin{table}[t]
    \caption{Performance of \proposedmodel{} w/o environmental exploration guidance and stable pattern preservation guidance on synthetic and real-world datasets. Bold numbers indicate the best performance.}
    \label{tab:ablation_results}  
  
    \begin{center}
    \resizebox{\columnwidth}{!}{
    \begin{tabular}{ccccccccc}
        \toprule
        \multirow{1}{*}{$\lambda$} & \multirow{1}{*}{$\alpha$} & \multicolumn{1}{c}{Motif-base} & \multicolumn{1}{c}{Molhiv-scaffold} & \multicolumn{1}{c}{GOOD-SST2-length} \\\midrule
        \multicolumn{2}{c}{ERM} & \multicolumn{1}{c}{$68.66 \pm 4.25$} & \multicolumn{1}{c}{$69.58 \pm 2.51$} & \multicolumn{1}{c}{$81.30 \pm 0.35$} &  \\
        \ding{55} & \ding{55} & \multicolumn{1}{c}{$68.55\pm 6.04$} & \multicolumn{1}{c}{$68.94 \pm 1.26$} & \multicolumn{1}{c}{$80.86 \pm 0.76$} &  \\
        \checkmark & \ding{55} & \multicolumn{1}{c}{$66.25\pm 7.42$} & \multicolumn{1}{c}{$70.01 \pm 1.71$} & \multicolumn{1}{c}{$78.87 \pm 3.04$} &  \\
        \ding{55} & \checkmark & \multicolumn{1}{c}{$74.57\pm 4.50$} & \multicolumn{1}{c}{$71.71 \pm 1.77$} & \multicolumn{1}{c}{$81.59 \pm 0.65$} &  \\
        \multicolumn{2}{c}{\proposedmodel{}} & $\mathbf{75.25 \pm 3.84}$ & $\mathbf{72.67 \pm 1.28}$ & $\mathbf{82.69 \pm 0.28}$ &  \\
         \bottomrule
    \end{tabular}
    }
    \end{center}

\end{table}

\noindent\textbf{Visualization of OOD graphs}
We further demonstrate the efficacy of \proposedmodel{} by visualizing the OOD graphs generated by our approach in Table~\ref{tab:ood_graphs_vis}. In this visualization, three label-determining motifs—house, cycle, and crane—are highlighted in red, while the three environmental base graphs in the training distribution—wheel, tree, and ladder—are indicated in green. As illustrated in Table~\ref{tab:ood_graphs_vis}, increasing $\lambda$ leads to gradual modifications in the structures of the base graphs, while the motifs remain preserved. 


\begin{table}

  \caption{Visualizations of the augmented GOOD-Motif-base graphs generated by \proposedmodel{}. The graphs $\tilde{G}$ with $\lambda=0.1$, $\lambda=0.2$, and $\lambda=0.3$ represent the OOD graphs generated by \proposedmodel{} under different values of $\lambda$.}
  \label{tab:ood_graphs_vis}  

  \begin{center}
    \begin{tabular}{cccc}
        \toprule
        Original Graphs & $\tilde{G} (\lambda=0.1)$ & $\tilde{G} (\lambda=0.2)$ & $\tilde{G} (\lambda=0.3)$\\
        \midrule
        \multicolumn{4}{c}{\textbf{House} Class} \\
        \includegraphics[width=10mm, height=10mm]{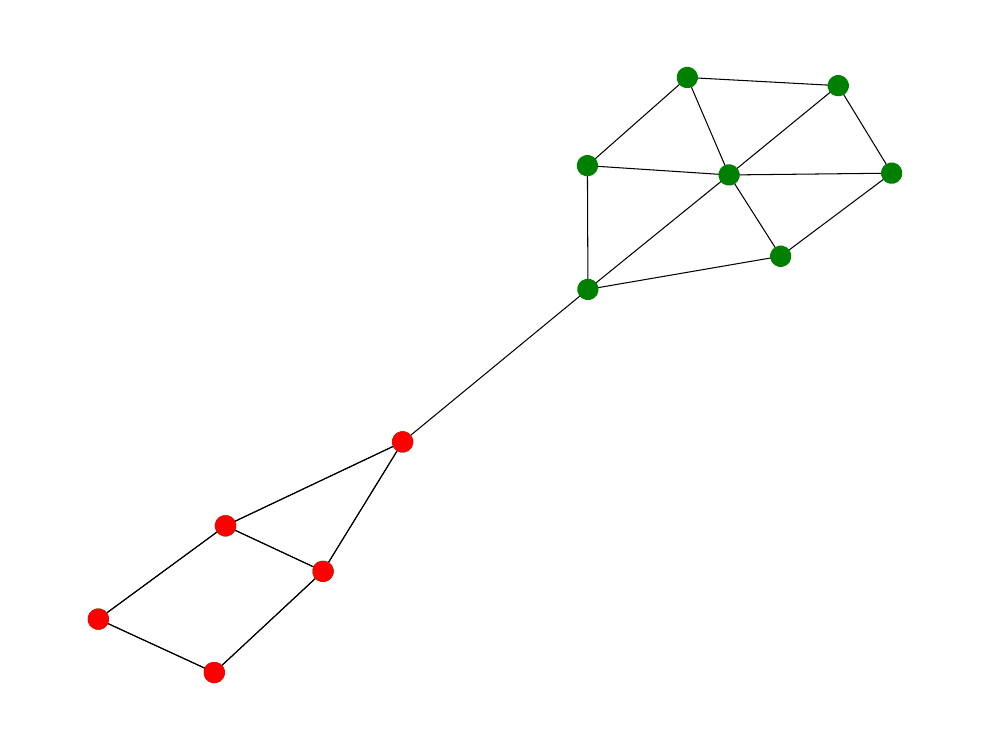} &
        \includegraphics[width=10mm, height=10mm]{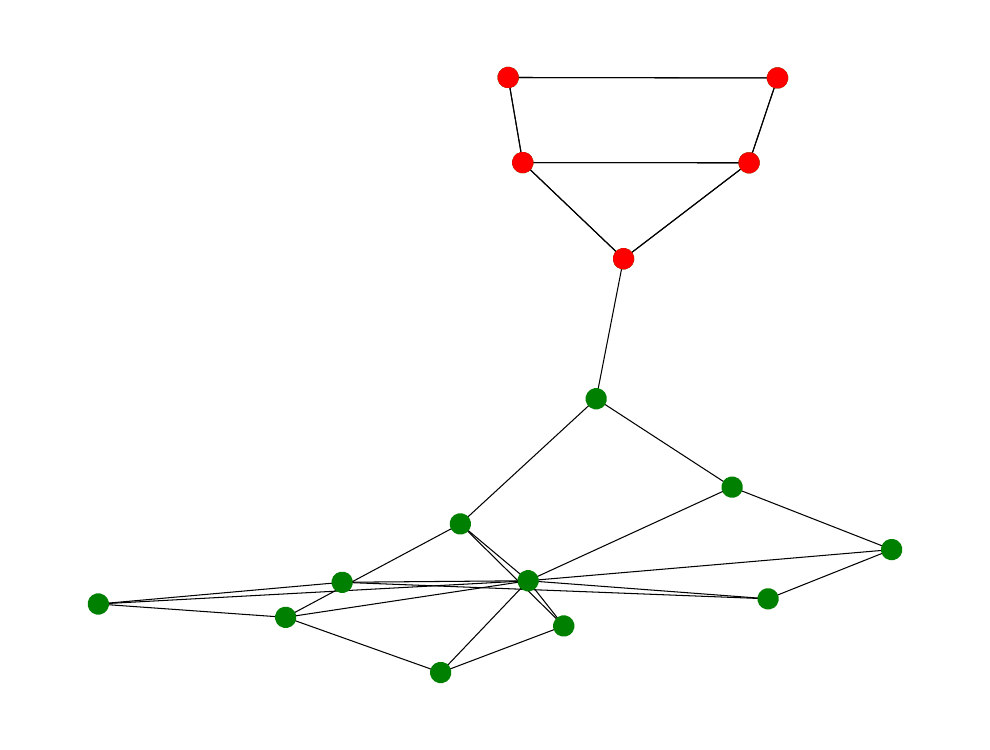} &
        \includegraphics[width=10mm, height=10mm]{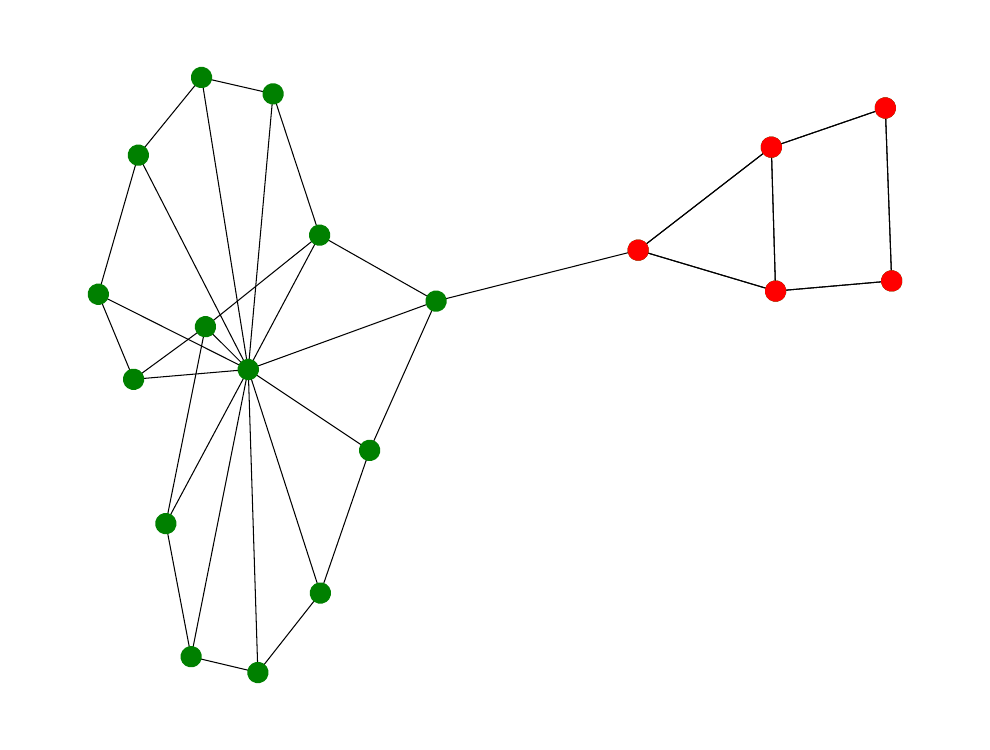} &
        \includegraphics[width=10mm, height=10mm]{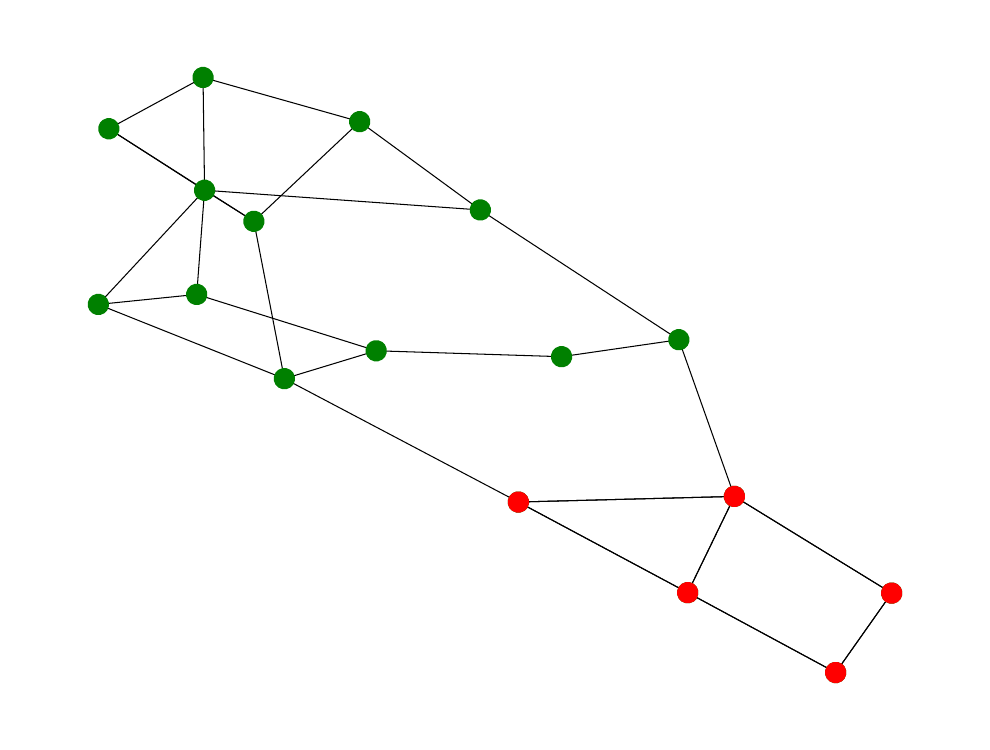} \\
        \includegraphics[width=10mm, height=10mm]{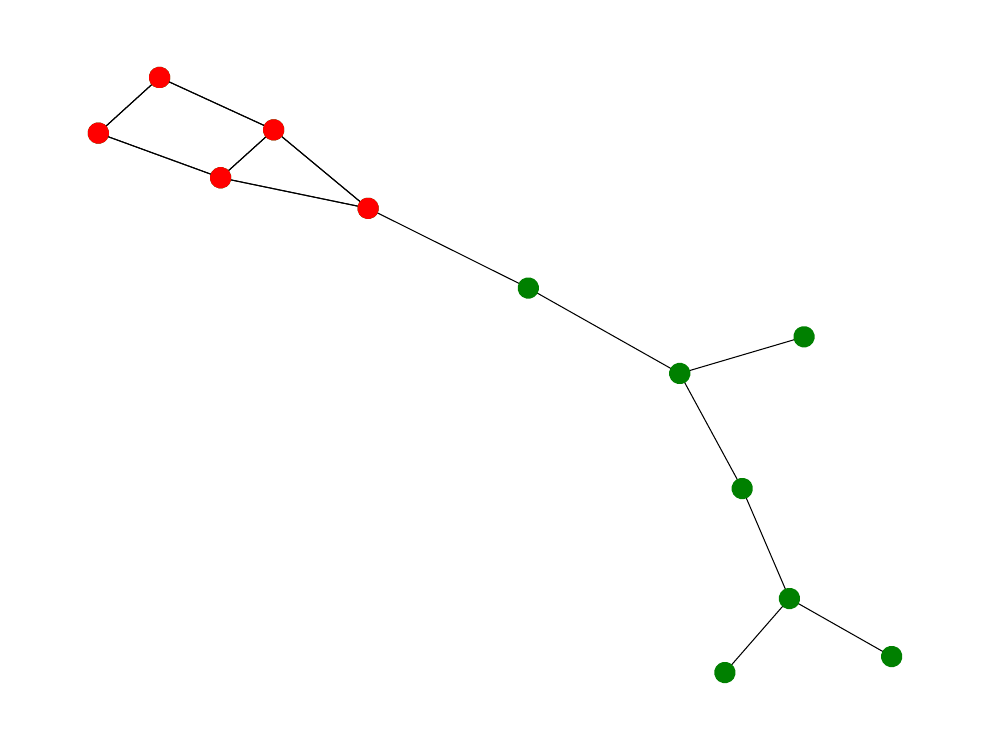} &
        \includegraphics[width=10mm, height=10mm]{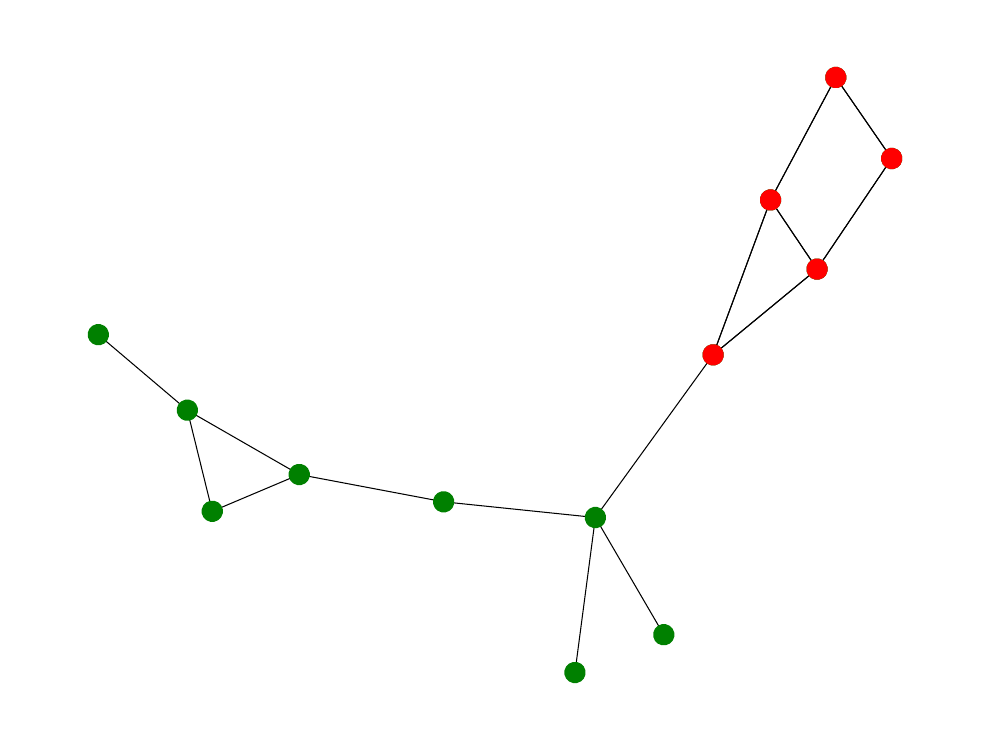} &
        \includegraphics[width=10mm, height=10mm]{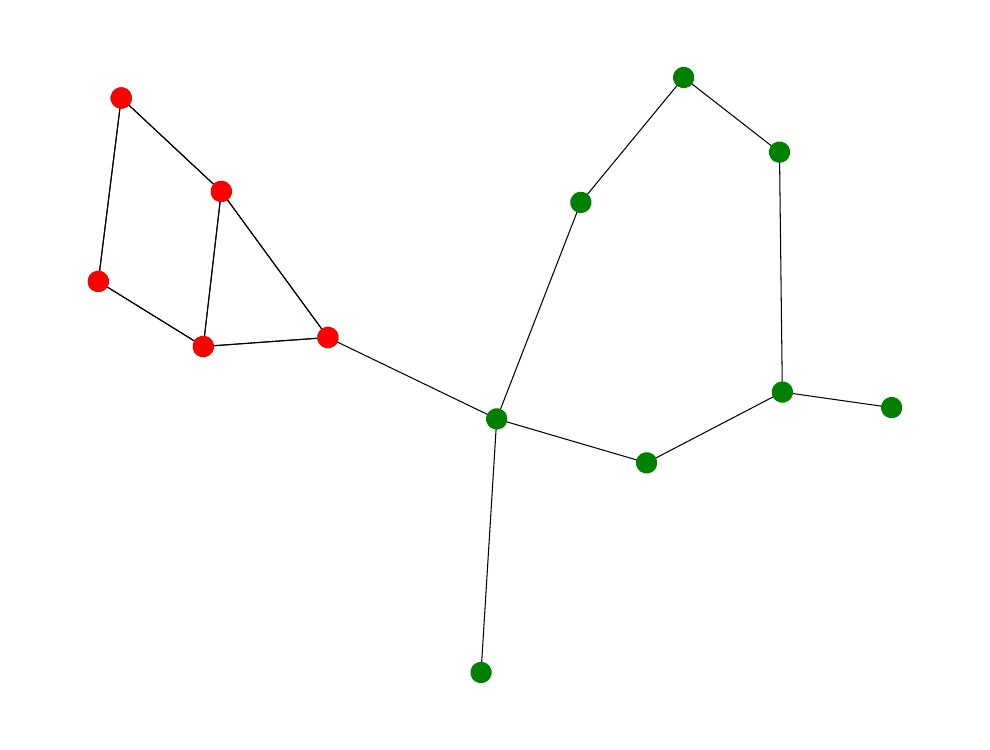} &
        \includegraphics[width=10mm, height=10mm]{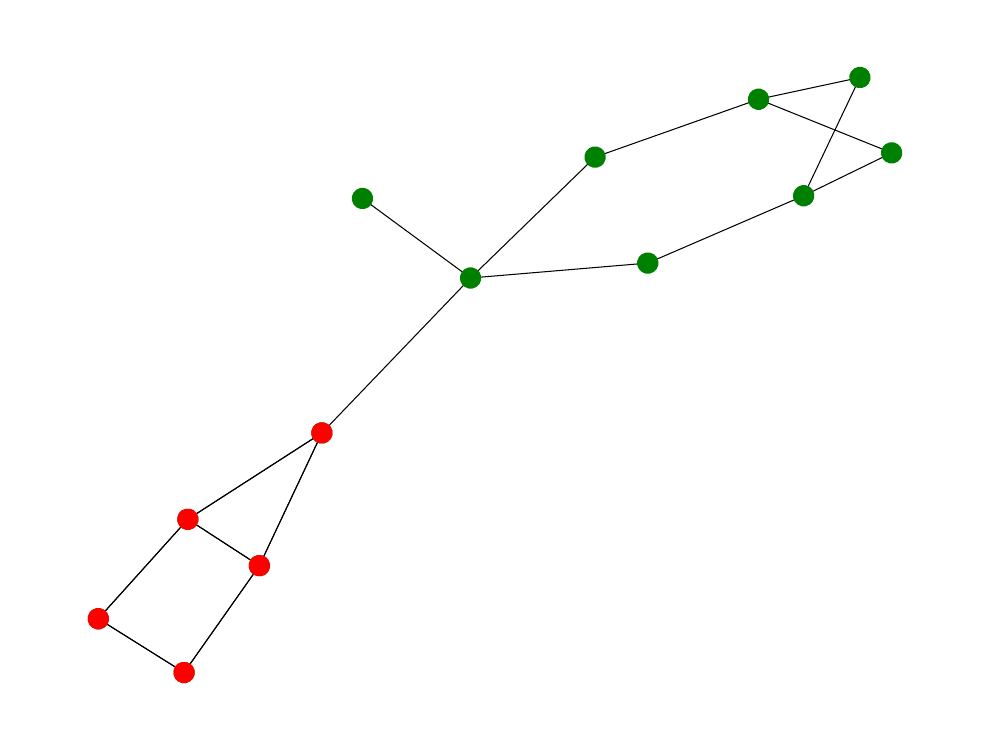} \\
        \midrule
        \multicolumn{4}{c}{\textbf{Cycle} Class} \\
        \includegraphics[width=10mm, height=10mm]{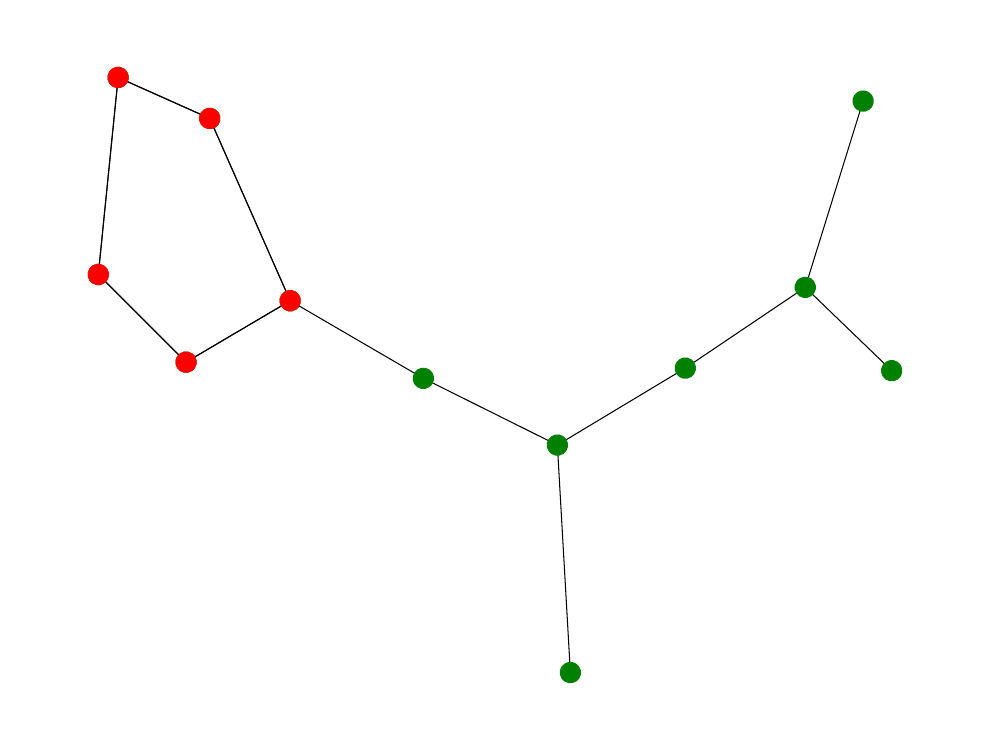} &
        \includegraphics[width=10mm, height=10mm]{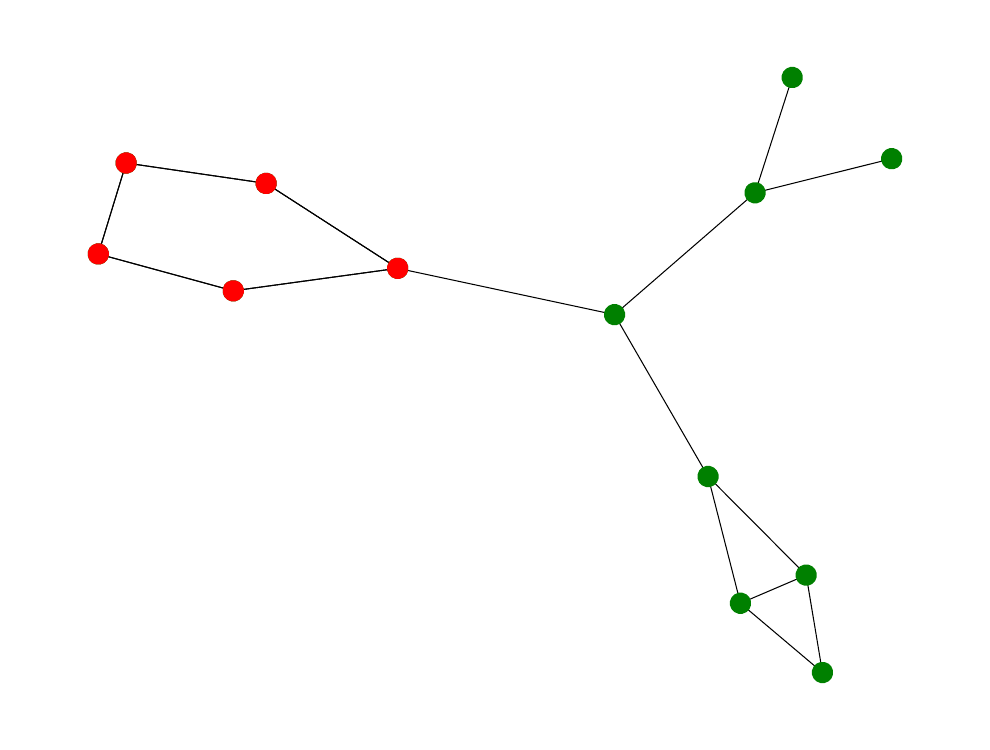} &
        \includegraphics[width=10mm, height=10mm]{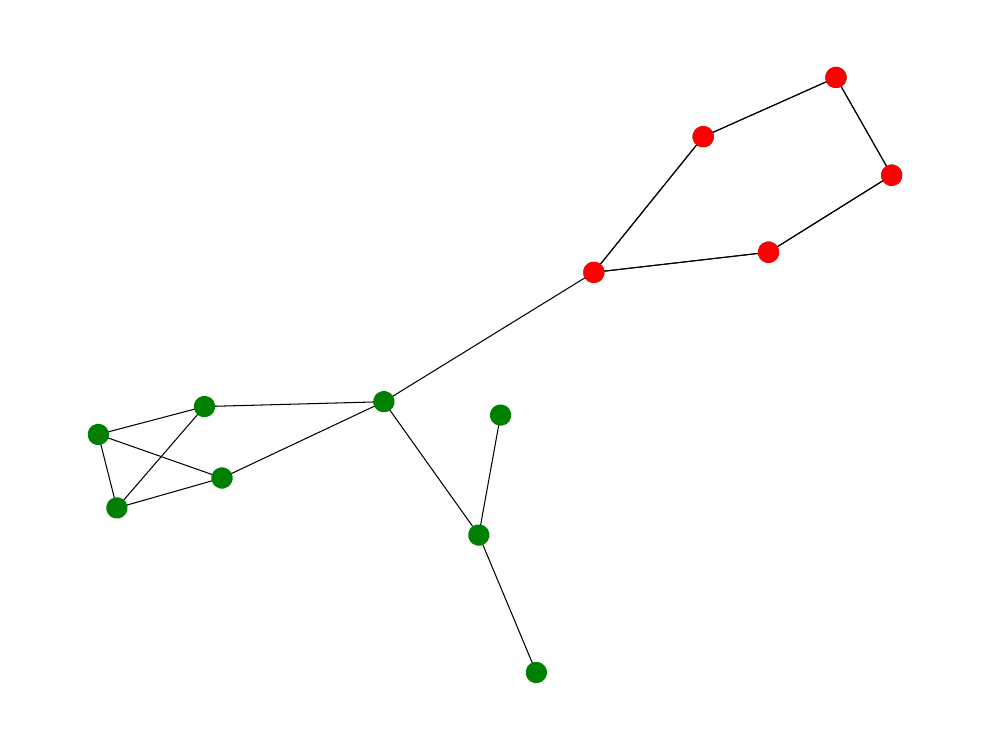} &
        \includegraphics[width=10mm, height=10mm]{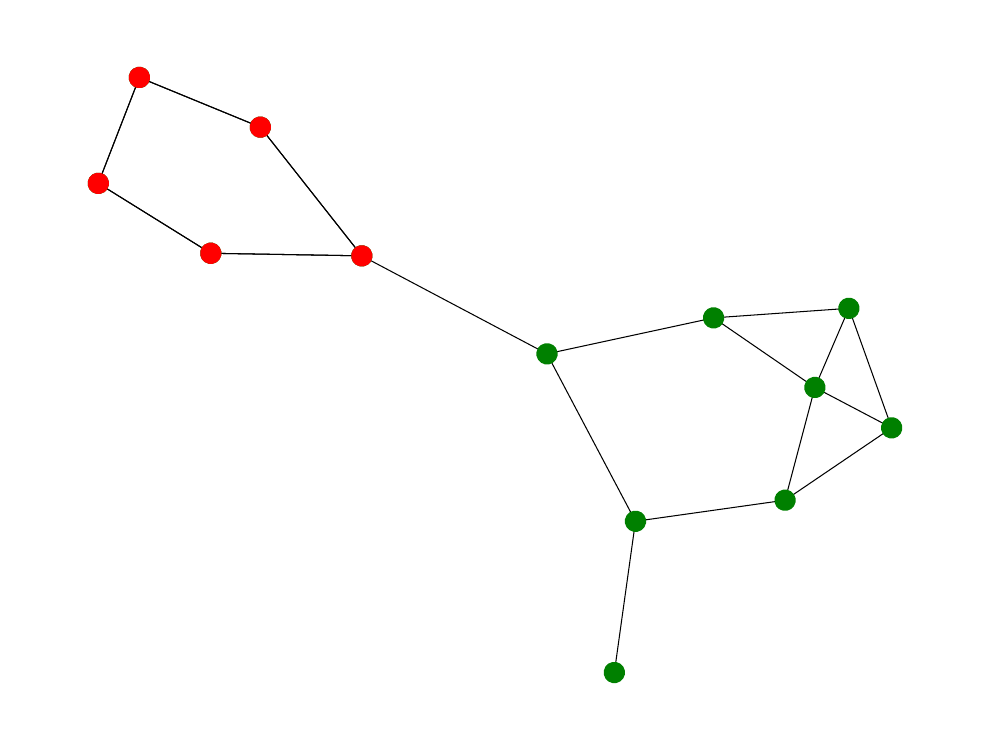} \\
        \includegraphics[width=10mm, height=10mm]{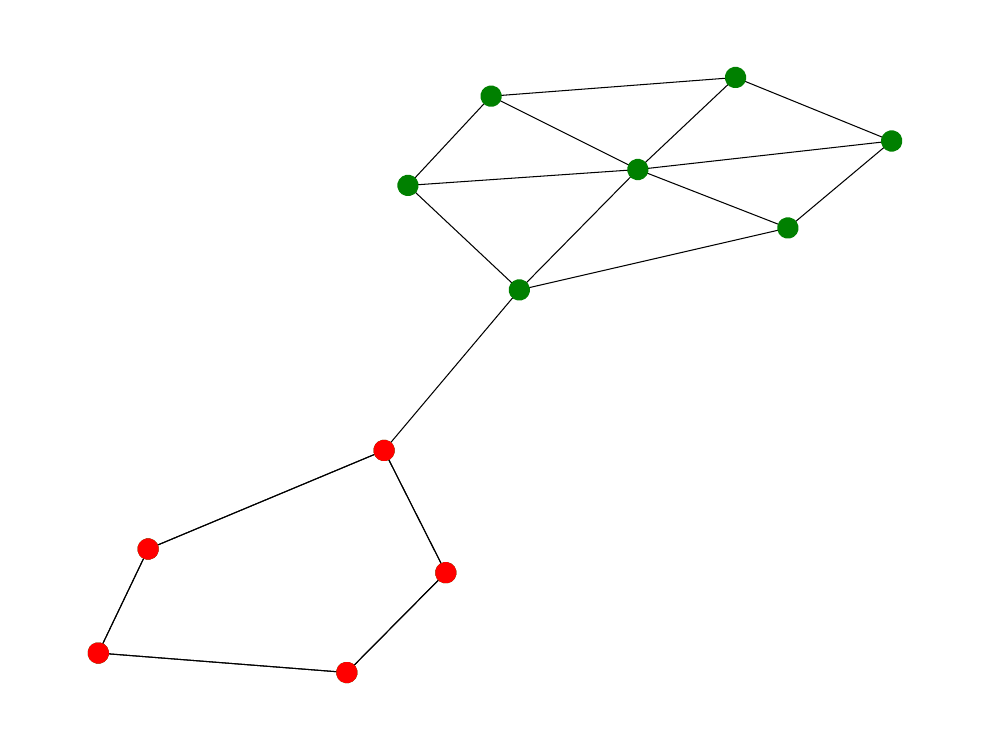} &
        \includegraphics[width=10mm, height=10mm]{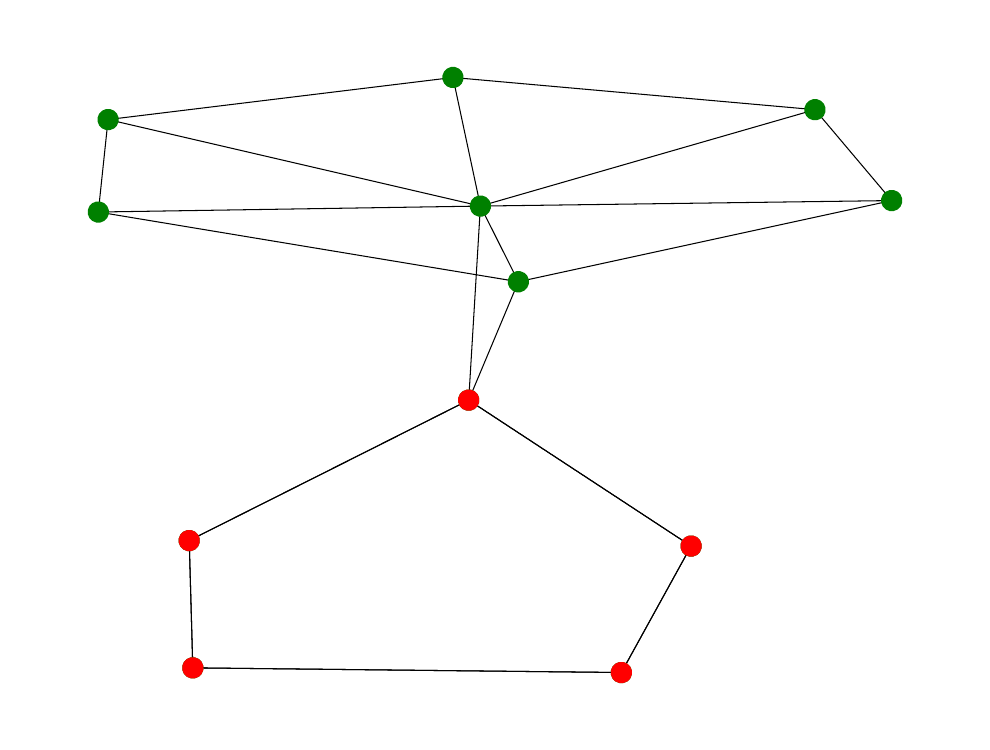} &
        \includegraphics[width=10mm, height=10mm]{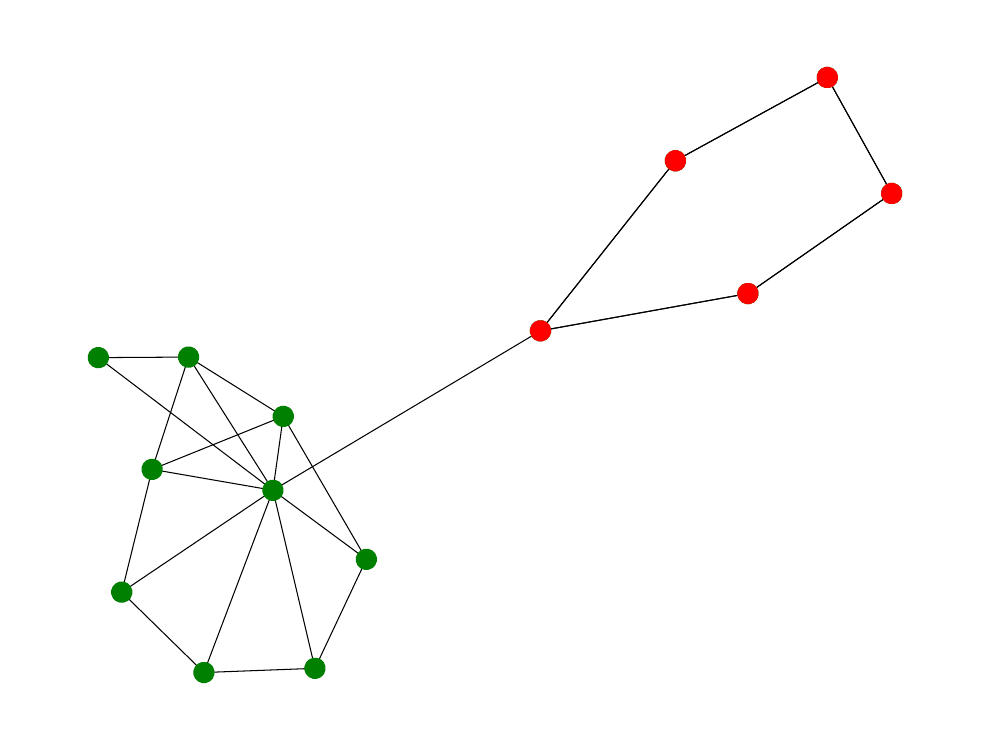} &
        \includegraphics[width=10mm, height=10mm]{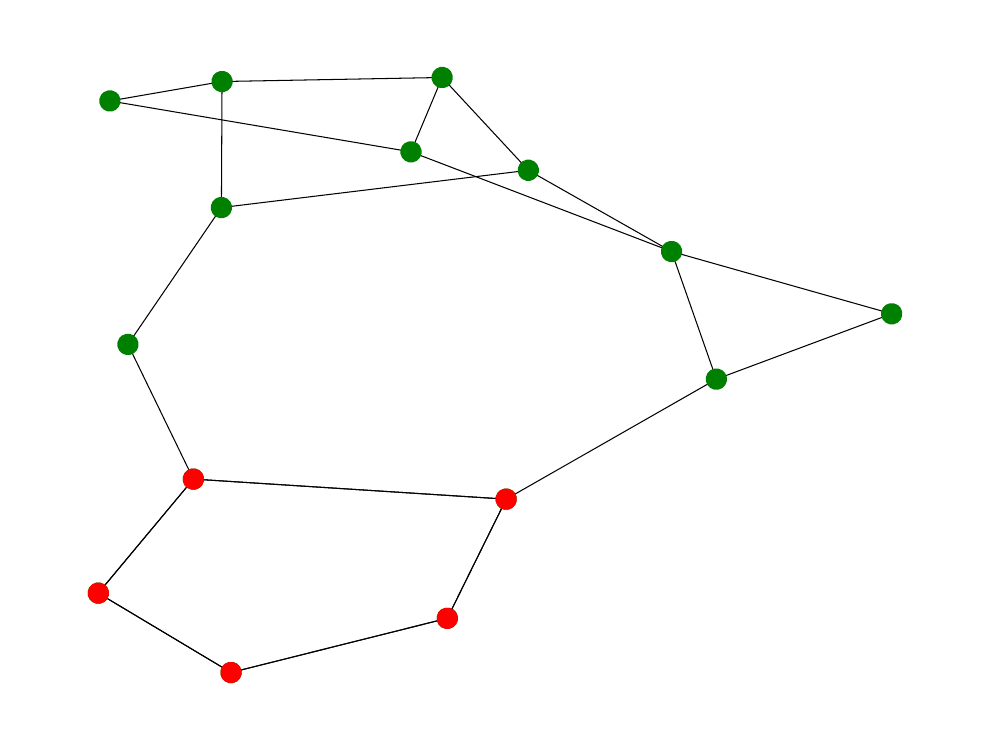} \\
        \midrule
        \multicolumn{4}{c}{\textbf{Crane} Class} \\
        \includegraphics[width=10mm, height=10mm]{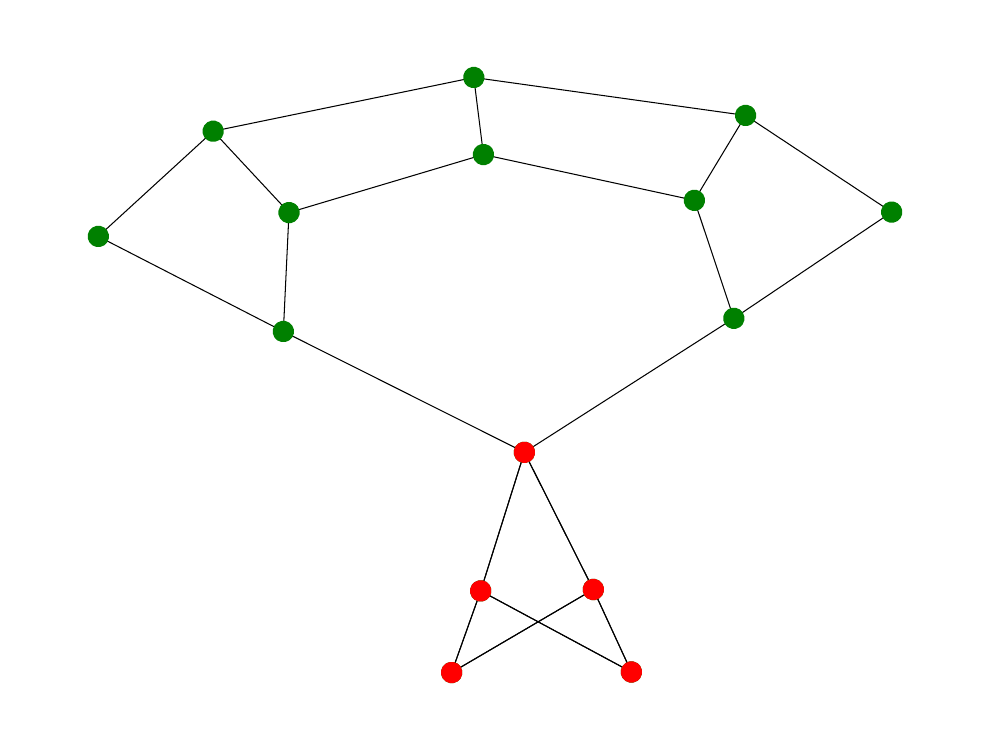} &
        \includegraphics[width=10mm, height=10mm]{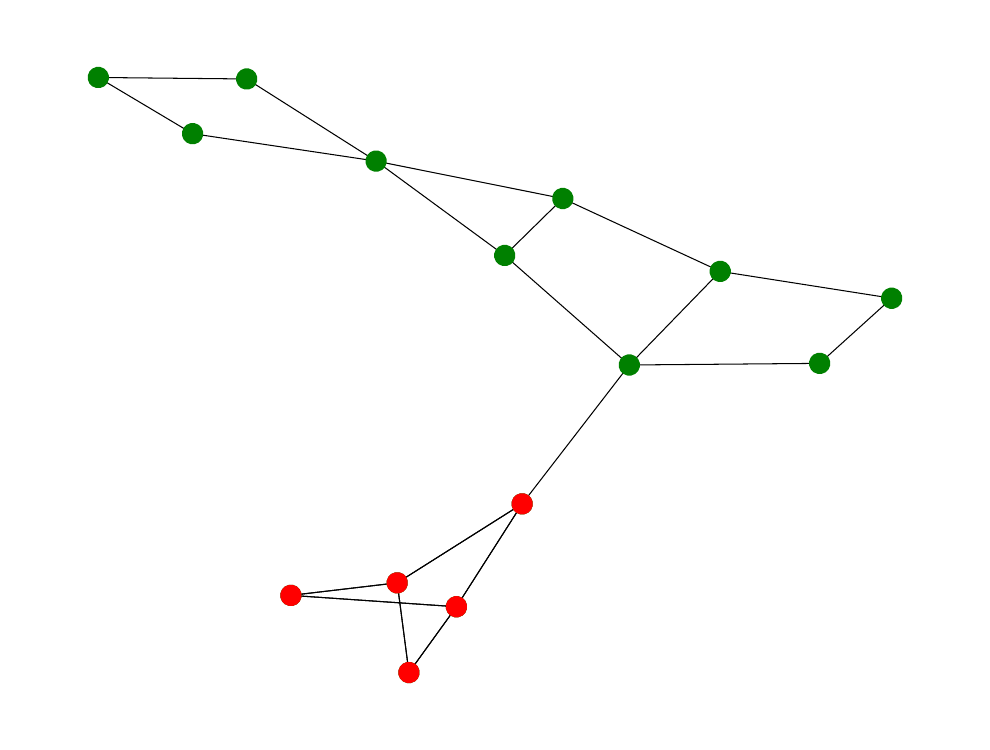} &
        \includegraphics[width=10mm, height=10mm]{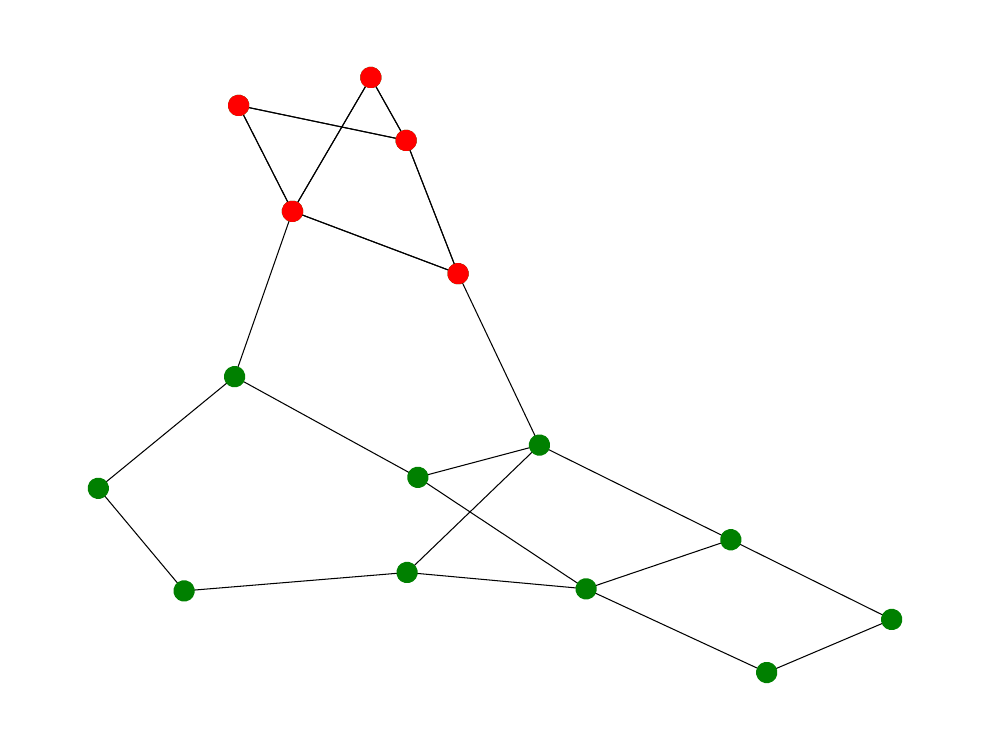} &
        \includegraphics[width=10mm, height=10mm]{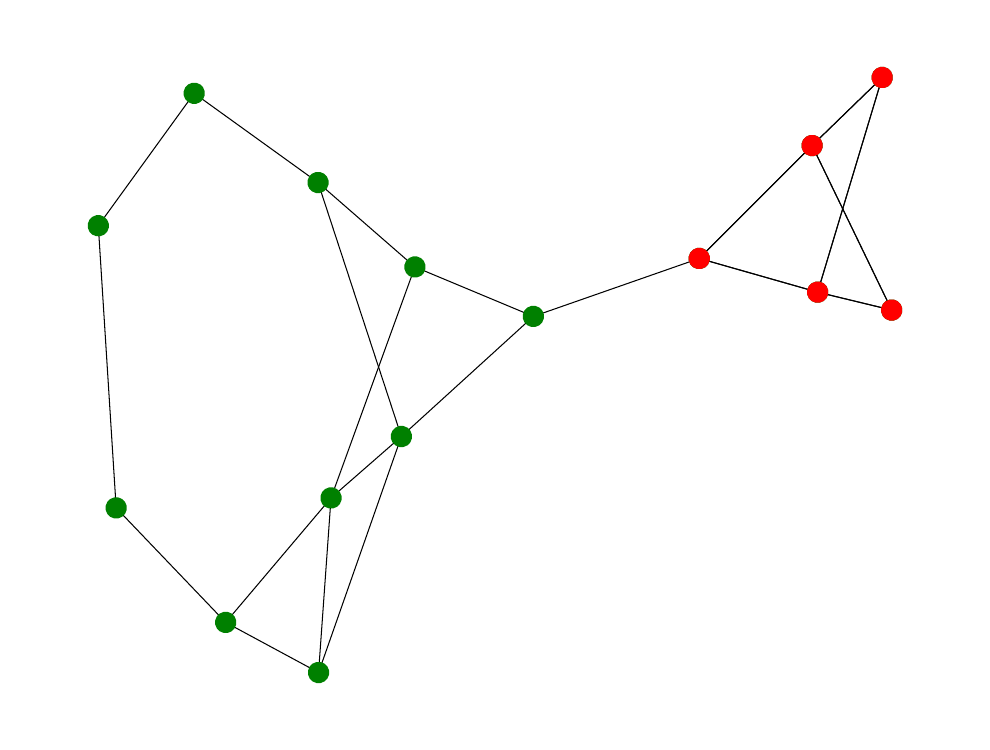} \\
        \includegraphics[width=10mm, height=10mm]{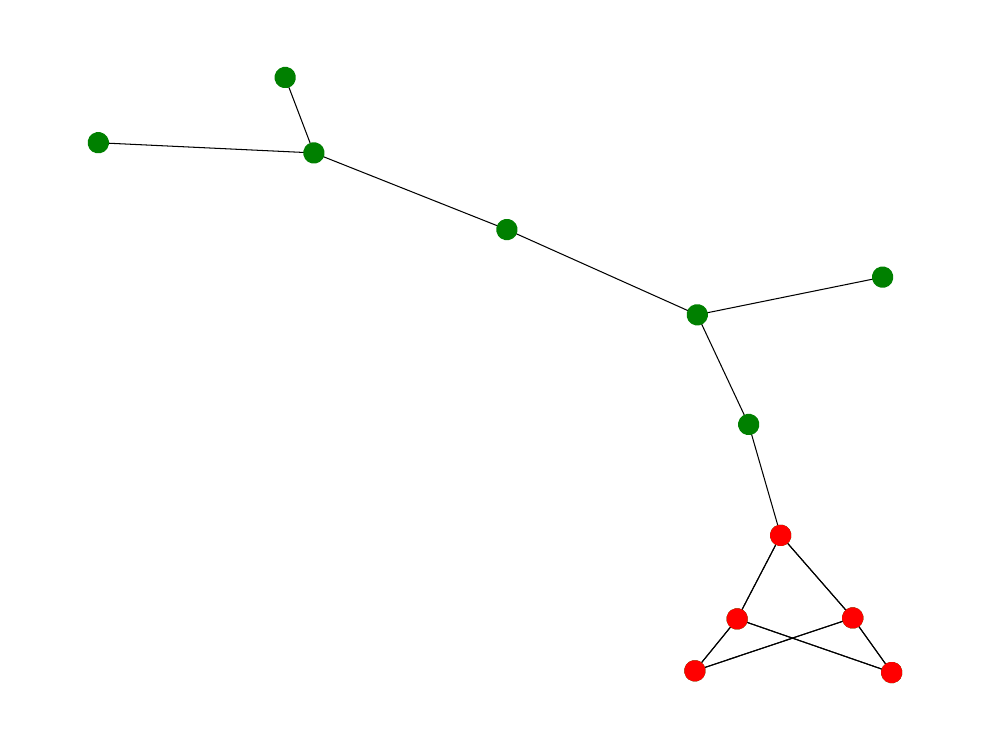} &
        \includegraphics[width=10mm, height=10mm]{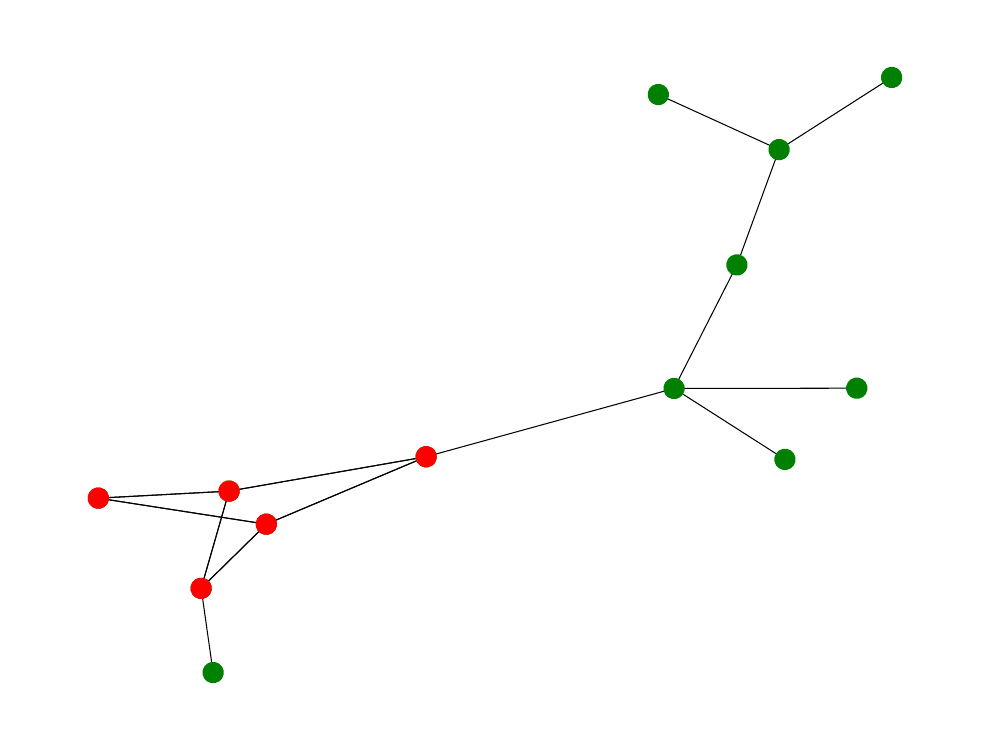} &
        \includegraphics[width=10mm, height=10mm]{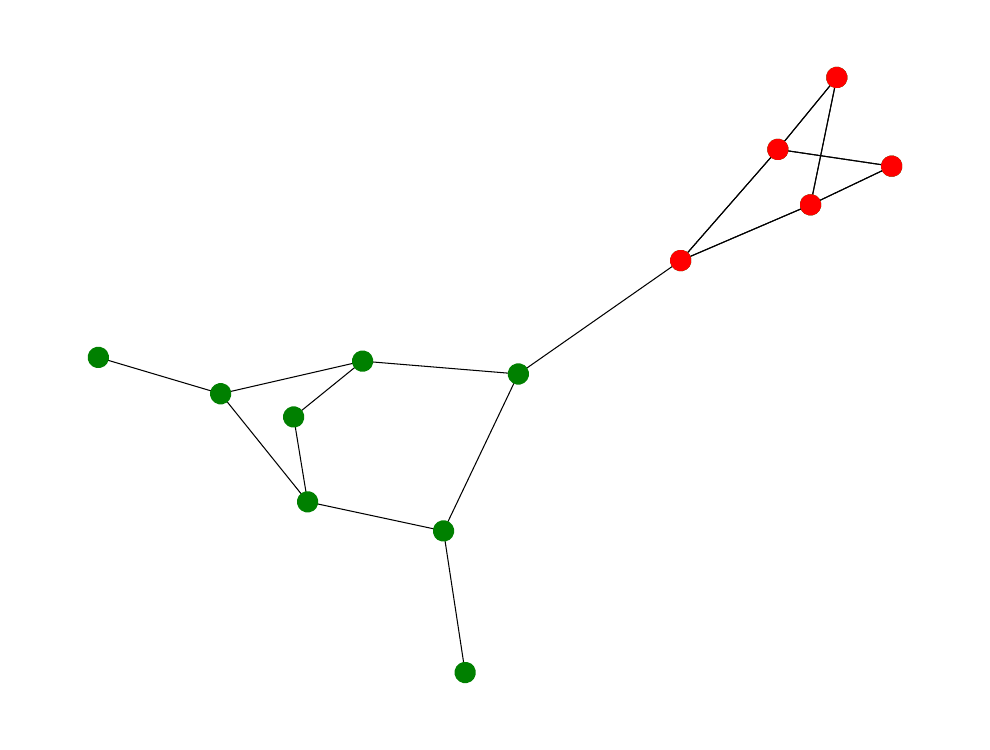} &
        \includegraphics[width=10mm, height=10mm]{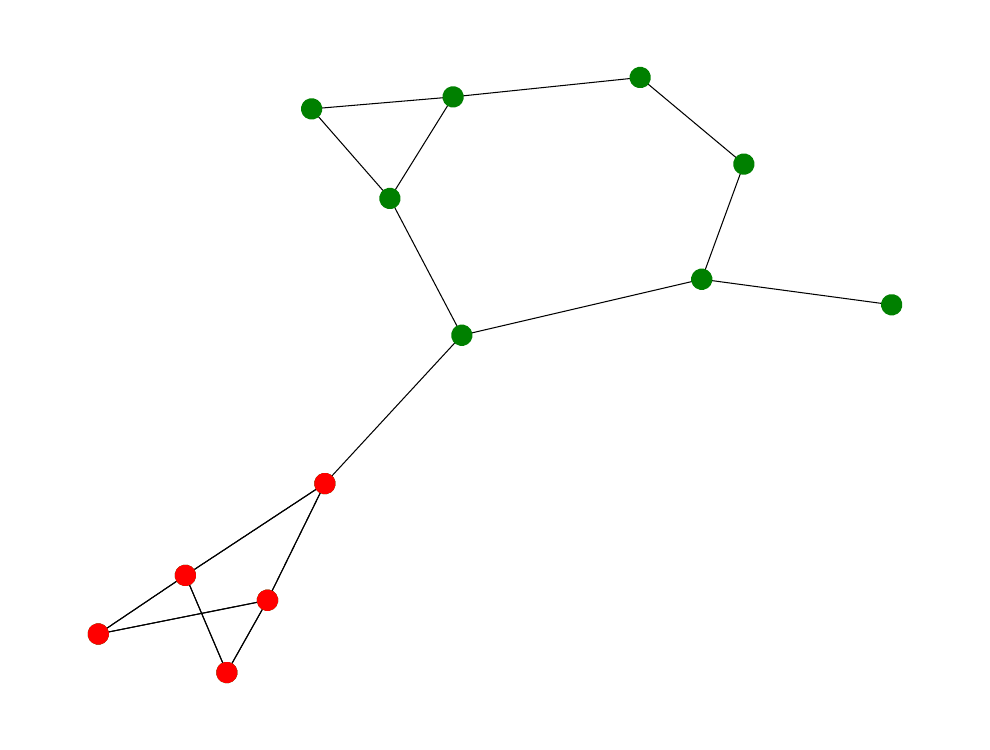} \\
        \bottomrule
    \end{tabular}
    \end{center}

\end{table}

\noindent\textbf{Visualization of OOD molecules}
We showcase the effectiveness of \proposedmodel{} by visualizing the OOD molecules generated by our approach in Table~\ref{tab:ood_mols_vis}. As illustrated in Table~\ref{tab:ood_mols_vis}, increasing $\lambda$ results in gradual modifications to the structures of the molecular graphs, while the properties of the molecules remain preserved. This controlled exploration allows \proposedmodel{} to produce chemically meaningful and diverse OOD molecules without disrupting their underlying validity. Notably, the changes remain smooth and interpretable across different $\lambda$ values, highlighting the tunable nature of our generation process.

\begin{table}

  \caption{Visualizations of the augmented GOOD-Molhiv-scaffold graphs generated by \proposedmodel{}. The graphs $\tilde{G}$ with $\lambda=0.1$, $\lambda=0.2$, and $\lambda=0.3$ represent the OOD graphs generated by \proposedmodel{} under different values of $\lambda$.}
  \label{tab:ood_mols_vis}  

  \centering

  \begingroup
  \setlength{\tabcolsep}{2pt}        
  \renewcommand{\arraystretch}{0.8} 
  \setlength{\aboverulesep}{0pt}
  \setlength{\belowrulesep}{0pt}
  \setlength{\abovetopsep}{0pt}
  \setlength{\belowbottomsep}{0pt}
    \begin{tabular}{cccc}
        \toprule
        Original Graphs & $\tilde{G} (\lambda=0.1)$ & $\tilde{G} (\lambda=0.2)$ & $\tilde{G} (\lambda=0.3)$\\ 
        \midrule

        \includegraphics[width=19mm, trim=10 50 10 50,clip]{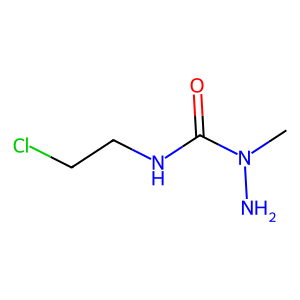} &
        \includegraphics[width=19mm, trim=10 50 10 50,clip]{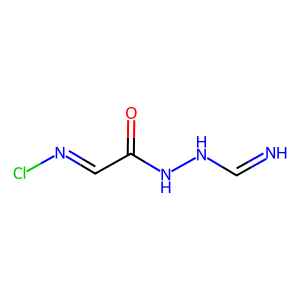} &
        \includegraphics[width=19mm, trim=10 50 10 50,clip]{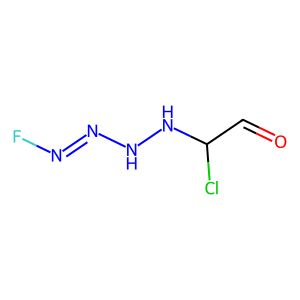} &
        \includegraphics[width=19mm, trim=10 50 10 50,clip]{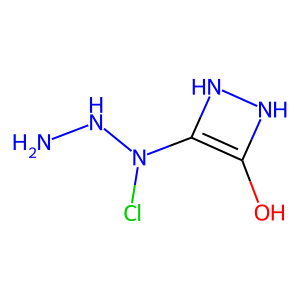} \\ [-1em]
        \includegraphics[width=19mm, trim=10 60 10 60,clip]{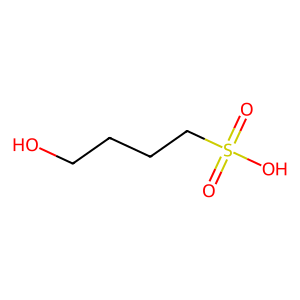} &
        \includegraphics[width=19mm, trim=10 50 10 50,clip]{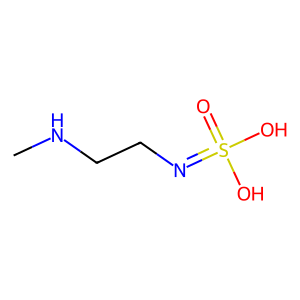} &
        \includegraphics[width=19mm, trim=10 50 10 50,clip]{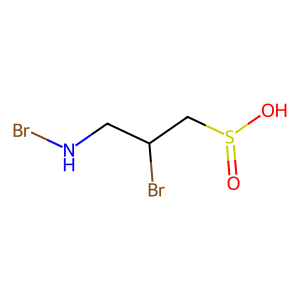} &
        \includegraphics[width=19mm, trim=10 60 10 60,clip]{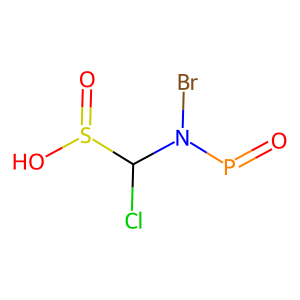} \\[-1em]
        \includegraphics[width=19mm, trim=10 50 10 50,clip]{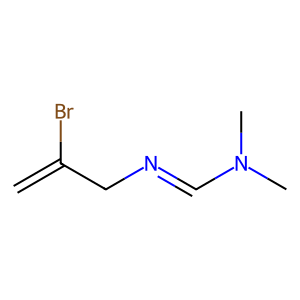} &
        \includegraphics[width=19mm, trim=10 50 10 50,clip]{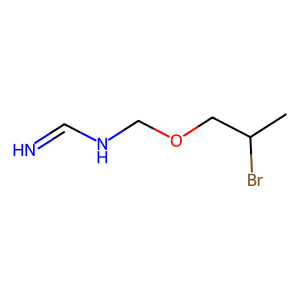} &
        \includegraphics[width=19mm, trim=10 50 10 50,clip]{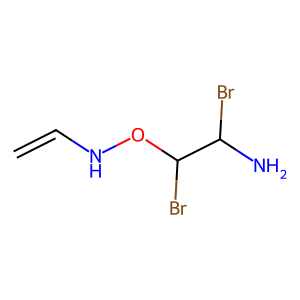} &
        \includegraphics[width=19mm, trim=10 50 10 80,clip]{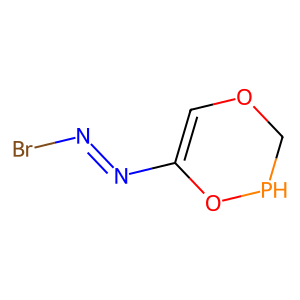} \\[-1em]
        \includegraphics[width=19mm, trim=10 50 10 50,clip]{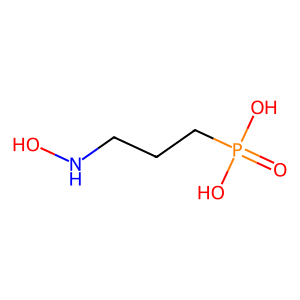} &
        \includegraphics[width=19mm, trim=10 50 10 50,clip]{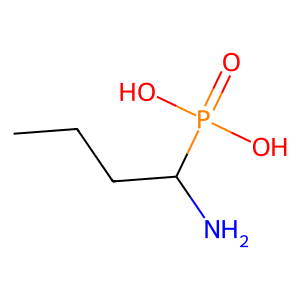} &
        \includegraphics[width=19mm, trim=10 50 10 50,clip]{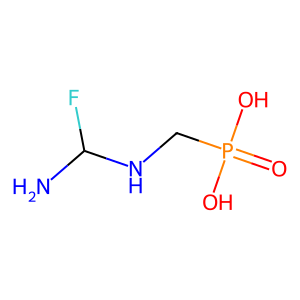} &
        \includegraphics[width=19mm, trim=10 50 10 50,clip]{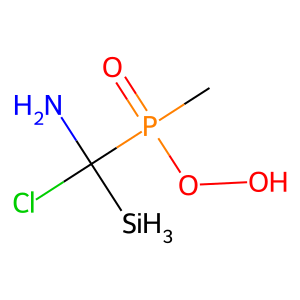} \\

        \bottomrule
    \end{tabular}
    \endgroup

\end{table}

\section{Conclusion}

We presented \proposedmodel{}, a score-based out-of-distribution graph generation framework for addressing covariate shift in graph learning. By coupling a conditional score with an exploration parameter, \proposedmodel{} explores low-density regions beyond the training support in a \emph{controlled} manner while preserving label-determining stable patterns. Unlike edit-based augmentations, our approach jointly synthesizes structures, node features, and edge features and does not require explicit decomposition into stable and environmental components. Across synthetic, semi-artificial, and real benchmarks from GOOD, \proposedmodel{} consistently improves OOD generalization over invariant-learning and augmentation baselines. The exploration knob~$\lambda$ yields monotonic increases in MMD while maintaining a high probability of retaining stable patterns, demonstrating both controllability and fidelity. Given that distribution shift is pervasive on the Web (e.g., evolving social, hyperlink, and interaction graphs), these results indicate practical value for robust web-scale modeling.

Overall, \proposedmodel{} provides a principled and controllable path to web-relevant graph augmentation, improving robustness to the distribution shifts that dominate modern web applications.

\bibliographystyle{ACM-Reference-Format}
\bibliography{sample-base}

\begin{thebibliography}{43}
\providecommand{\natexlab}[1]{#1}
\providecommand{\url}[1]{\texttt{#1}}
\expandafter\ifx\csname urlstyle\endcsname\relax
  \providecommand{\doi}[1]{doi: #1}\else
  \providecommand{\doi}{doi: \begingroup \urlstyle{rm}\Url}\fi

\bibitem[Ahuja et~al.(2021)Ahuja, Caballero, Zhang, Gagnon-Audet, Bengio, Mitliagkas, and Rish]{ahuja2021invariance}
Kartik Ahuja, Ethan Caballero, Dinghuai Zhang, Jean-Christophe Gagnon-Audet, Yoshua Bengio, Ioannis Mitliagkas, and Irina Rish.
\newblock Invariance principle meets information bottleneck for out-of-distribution generalization.
\newblock \emph{Advances in Neural Information Processing Systems}, 34:\penalty0 3438--3450, 2021.

\bibitem[Arjovsky et~al.(2019)Arjovsky, Bottou, Gulrajani, and Lopez-Paz]{arjovsky2019invariant}
Martin Arjovsky, L{\'e}on Bottou, Ishaan Gulrajani, and David Lopez-Paz.
\newblock Invariant risk minimization.
\newblock \emph{arXiv preprint arXiv:1907.02893}, 2019.

\bibitem[Bazhenov et~al.(2022)Bazhenov, Ivanov, Panov, Zaytsev, and Burnaev]{bazhenov2022towards}
Gleb Bazhenov, Sergei Ivanov, Maxim Panov, Alexey Zaytsev, and Evgeny Burnaev.
\newblock Towards ood detection in graph classification from uncertainty estimation perspective.
\newblock \emph{arXiv preprint arXiv:2206.10691}, 2022.

\bibitem[Bemis \& Murcko(1996)Bemis and Murcko]{bemis1996properties}
Guy~W Bemis and Mark~A Murcko.
\newblock The properties of known drugs. 1. molecular frameworks.
\newblock \emph{Journal of medicinal chemistry}, 39\penalty0 (15):\penalty0 2887--2893, 1996.

\bibitem[Chen et~al.(2022)Chen, Zhang, Bian, Yang, Kaili, Xie, Liu, Han, and Cheng]{chen2022learning}
Yongqiang Chen, Yonggang Zhang, Yatao Bian, Han Yang, MA~Kaili, Binghui Xie, Tongliang Liu, Bo~Han, and James Cheng.
\newblock Learning causally invariant representations for out-of-distribution generalization on graphs.
\newblock \emph{Advances in Neural Information Processing Systems}, 35:\penalty0 22131--22148, 2022.

\bibitem[Chen et~al.(2024)Chen, Bian, Zhou, Xie, Han, and Cheng]{chen2024does}
Yongqiang Chen, Yatao Bian, Kaiwen Zhou, Binghui Xie, Bo~Han, and James Cheng.
\newblock Does invariant graph learning via environment augmentation learn invariance?
\newblock \emph{Advances in Neural Information Processing Systems}, 36, 2024.

\bibitem[Dwivedi \& Bresson(2020)Dwivedi and Bresson]{dwivedi2020generalization}
Vijay~Prakash Dwivedi and Xavier Bresson.
\newblock A generalization of transformer networks to graphs.
\newblock \emph{arXiv preprint arXiv:2012.09699}, 2020.

\bibitem[Feng et~al.(2020)Feng, Zhang, Dong, Han, Luan, Xu, Yang, Kharlamov, and Tang]{feng2020graph}
Wenzheng Feng, Jie Zhang, Yuxiao Dong, Yu~Han, Huanbo Luan, Qian Xu, Qiang Yang, Evgeny Kharlamov, and Jie Tang.
\newblock Graph random neural networks for semi-supervised learning on graphs.
\newblock \emph{Advances in neural information processing systems}, 33:\penalty0 22092--22103, 2020.

\bibitem[Ganin et~al.(2016)Ganin, Ustinova, Ajakan, Germain, Larochelle, Laviolette, March, and Lempitsky]{ganin2016domain}
Yaroslav Ganin, Evgeniya Ustinova, Hana Ajakan, Pascal Germain, Hugo Larochelle, Fran{\c{c}}ois Laviolette, Mario March, and Victor Lempitsky.
\newblock Domain-adversarial training of neural networks.
\newblock \emph{Journal of machine learning research}, 17\penalty0 (59):\penalty0 1--35, 2016.

\bibitem[Gilmer et~al.(2017)Gilmer, Schoenholz, Riley, Vinyals, and Dahl]{gilmer2017neural}
Justin Gilmer, Samuel~S Schoenholz, Patrick~F Riley, Oriol Vinyals, and George~E Dahl.
\newblock Neural message passing for quantum chemistry.
\newblock In \emph{International conference on machine learning}, pp.\  1263--1272. PMLR, 2017.

\bibitem[Gui et~al.(2022)Gui, Li, Wang, and Ji]{gui2022good}
Shurui Gui, Xiner Li, Limei Wang, and Shuiwang Ji.
\newblock Good: A graph out-of-distribution benchmark.
\newblock \emph{Advances in Neural Information Processing Systems}, 35:\penalty0 2059--2073, 2022.

\bibitem[Han et~al.(2022)Han, Jiang, Liu, and Hu]{han2022g}
Xiaotian Han, Zhimeng Jiang, Ninghao Liu, and Xia Hu.
\newblock G-mixup: Graph data augmentation for graph classification.
\newblock In \emph{International Conference on Machine Learning}, pp.\  8230--8248. PMLR, 2022.

\bibitem[Huang et~al.(2024)Huang, Yang, Zhou, and Yan]{huang2024enhancing}
Zheng Huang, Qihui Yang, Dawei Zhou, and Yujun Yan.
\newblock Enhancing size generalization in graph neural networks through disentangled representation learning.
\newblock \emph{arXiv preprint arXiv:2406.04601}, 2024.

\bibitem[Jo et~al.(2022)Jo, Lee, and Hwang]{jo2022score}
Jaehyeong Jo, Seul Lee, and Sung~Ju Hwang.
\newblock Score-based generative modeling of graphs via the system of stochastic differential equations.
\newblock In \emph{International Conference on Machine Learning}, pp.\  10362--10383. PMLR, 2022.

\bibitem[Kong et~al.(2022)Kong, Li, Ding, Wu, Zhu, Ghanem, Taylor, and Goldstein]{kong2022robust}
Kezhi Kong, Guohao Li, Mucong Ding, Zuxuan Wu, Chen Zhu, Bernard Ghanem, Gavin Taylor, and Tom Goldstein.
\newblock Robust optimization as data augmentation for large-scale graphs.
\newblock In \emph{Proceedings of the IEEE/CVF Conference on Computer Vision and Pattern Recognition}, pp.\  60--69, 2022.

\bibitem[Krueger et~al.(2021)Krueger, Caballero, Jacobsen, Zhang, Binas, Zhang, Le~Priol, and Courville]{krueger2021out}
David Krueger, Ethan Caballero, Joern-Henrik Jacobsen, Amy Zhang, Jonathan Binas, Dinghuai Zhang, Remi Le~Priol, and Aaron Courville.
\newblock Out-of-distribution generalization via risk extrapolation (rex).
\newblock In \emph{International Conference on Machine Learning}, pp.\  5815--5826. PMLR, 2021.

\bibitem[Landrum et~al.(2016)]{landrum2016rdkit}
Greg Landrum et~al.
\newblock Rdkit: Open-source cheminformatics software. 2016.
\newblock \emph{URL http://www. rdkit. org/, https://github. com/rdkit/rdkit}, 149\penalty0 (150):\penalty0 650, 2016.

\bibitem[Lee et~al.(2023)Lee, Jo, and Hwang]{lee2023exploring}
Seul Lee, Jaehyeong Jo, and Sung~Ju Hwang.
\newblock Exploring chemical space with score-based out-of-distribution generation.
\newblock In \emph{International Conference on Machine Learning}, pp.\  18872--18892. PMLR, 2023.

\bibitem[Li et~al.(2024)Li, Gui, Luo, and Ji]{ligraph}
Xiner Li, Shurui Gui, Youzhi Luo, and Shuiwang Ji.
\newblock Graph structure extrapolation for out-of-distribution generalization.
\newblock In \emph{Forty-first International Conference on Machine Learning}, 2024.

\bibitem[Martinkus et~al.(2022)Martinkus, Loukas, Perraudin, and Wattenhofer]{martinkus2022spectre}
Karolis Martinkus, Andreas Loukas, Nathana{\"e}l Perraudin, and Roger Wattenhofer.
\newblock Spectre: Spectral conditioning helps to overcome the expressivity limits of one-shot graph generators.
\newblock In \emph{International Conference on Machine Learning}, pp.\  15159--15179. PMLR, 2022.

\bibitem[Miao et~al.(2022)Miao, Liu, and Li]{miao2022interpretable}
Siqi Miao, Mia Liu, and Pan Li.
\newblock Interpretable and generalizable graph learning via stochastic attention mechanism.
\newblock In \emph{International Conference on Machine Learning}, pp.\  15524--15543. PMLR, 2022.

\bibitem[Monti et~al.(2017)Monti, Boscaini, Masci, Rodola, Svoboda, and Bronstein]{monti2017geometric}
Federico Monti, Davide Boscaini, Jonathan Masci, Emanuele Rodola, Jan Svoboda, and Michael~M Bronstein.
\newblock Geometric deep learning on graphs and manifolds using mixture model cnns.
\newblock In \emph{Proceedings of the IEEE conference on computer vision and pattern recognition}, pp.\  5115--5124, 2017.

\bibitem[Preuer et~al.(2018)Preuer, Renz, Unterthiner, Hochreiter, and Klambauer]{preuer2018frechet}
Kristina Preuer, Philipp Renz, Thomas Unterthiner, Sepp Hochreiter, and Gunter Klambauer.
\newblock Fr{\'e}chet chemnet distance: a metric for generative models for molecules in drug discovery.
\newblock \emph{Journal of chemical information and modeling}, 58\penalty0 (9):\penalty0 1736--1741, 2018.

\bibitem[Rong et~al.(2019)Rong, Huang, Xu, and Huang]{rong2019dropedge}
Yu~Rong, Wenbing Huang, Tingyang Xu, and Junzhou Huang.
\newblock Dropedge: Towards deep graph convolutional networks on node classification.
\newblock \emph{arXiv preprint arXiv:1907.10903}, 2019.

\bibitem[Rosenfeld et~al.(2020)Rosenfeld, Ravikumar, and Risteski]{rosenfeld2020risks}
Elan Rosenfeld, Pradeep Ravikumar, and Andrej Risteski.
\newblock The risks of invariant risk minimization.
\newblock \emph{arXiv preprint arXiv:2010.05761}, 2020.

\bibitem[Sagawa et~al.(2019)Sagawa, Koh, Hashimoto, and Liang]{sagawa2019distributionally}
Shiori Sagawa, Pang~Wei Koh, Tatsunori~B Hashimoto, and Percy Liang.
\newblock Distributionally robust neural networks for group shifts: On the importance of regularization for worst-case generalization.
\newblock \emph{arXiv preprint arXiv:1911.08731}, 2019.

\bibitem[Song et~al.(2020)Song, Sohl-Dickstein, Kingma, Kumar, Ermon, and Poole]{song2020score}
Yang Song, Jascha Sohl-Dickstein, Diederik~P Kingma, Abhishek Kumar, Stefano Ermon, and Ben Poole.
\newblock Score-based generative modeling through stochastic differential equations.
\newblock \emph{arXiv preprint arXiv:2011.13456}, 2020.

\bibitem[Sui et~al.(2024)Sui, Wu, Wu, Cui, Li, Zhou, Wang, and He]{sui2024unleashing}
Yongduo Sui, Qitian Wu, Jiancan Wu, Qing Cui, Longfei Li, Jun Zhou, Xiang Wang, and Xiangnan He.
\newblock Unleashing the power of graph data augmentation on covariate distribution shift.
\newblock \emph{Advances in Neural Information Processing Systems}, 36, 2024.

\bibitem[Sun \& Saenko(2016)Sun and Saenko]{sun2016deep}
Baochen Sun and Kate Saenko.
\newblock Deep coral: Correlation alignment for deep domain adaptation.
\newblock In \emph{Computer Vision--ECCV 2016 Workshops: Amsterdam, The Netherlands, October 8-10 and 15-16, 2016, Proceedings, Part III 14}, pp.\  443--450. Springer, 2016.

\bibitem[Thakoor et~al.(2021)Thakoor, Tallec, Azar, Azabou, Dyer, Munos, Veli{\v{c}}kovi{\'c}, and Valko]{thakoor2021large}
Shantanu Thakoor, Corentin Tallec, Mohammad~Gheshlaghi Azar, Mehdi Azabou, Eva~L Dyer, Remi Munos, Petar Veli{\v{c}}kovi{\'c}, and Michal Valko.
\newblock Large-scale representation learning on graphs via bootstrapping.
\newblock \emph{arXiv preprint arXiv:2102.06514}, 2021.

\bibitem[Thompson et~al.(2022)Thompson, Knyazev, Ghalebi, Kim, and Taylor]{thompson2022on}
Rylee Thompson, Boris Knyazev, Elahe Ghalebi, Jungtaek Kim, and Graham~W. Taylor.
\newblock On evaluation metrics for graph generative models.
\newblock In \emph{International Conference on Learning Representations}, 2022.
\newblock URL \url{https://openreview.net/forum?id=EnwCZixjSh}.

\bibitem[Vignac et~al.(2022)Vignac, Krawczuk, Siraudin, Wang, Cevher, and Frossard]{vignac2022digress}
Clement Vignac, Igor Krawczuk, Antoine Siraudin, Bohan Wang, Volkan Cevher, and Pascal Frossard.
\newblock Digress: Discrete denoising diffusion for graph generation.
\newblock \emph{arXiv preprint arXiv:2209.14734}, 2022.

\bibitem[Wang et~al.(2021)Wang, Wang, Liang, Cai, and Hooi]{wang2021mixup}
Yiwei Wang, Wei Wang, Yuxuan Liang, Yujun Cai, and Bryan Hooi.
\newblock Mixup for node and graph classification.
\newblock In \emph{Proceedings of the Web Conference 2021}, pp.\  3663--3674, 2021.

\bibitem[Wu et~al.(2023)Wu, Chen, Yang, and Yan]{wu2023energy}
Qitian Wu, Yiting Chen, Chenxiao Yang, and Junchi Yan.
\newblock Energy-based out-of-distribution detection for graph neural networks.
\newblock \emph{arXiv preprint arXiv:2302.02914}, 2023.

\bibitem[Wu et~al.(2022)Wu, Wang, Zhang, He, and Chua]{wu2022discovering}
Ying-Xin Wu, Xiang Wang, An~Zhang, Xiangnan He, and Tat-Seng Chua.
\newblock Discovering invariant rationales for graph neural networks.
\newblock \emph{arXiv preprint arXiv:2201.12872}, 2022.

\bibitem[Wu et~al.(2018)Wu, Ramsundar, Feinberg, Gomes, Geniesse, Pappu, Leswing, and Pande]{wu2018moleculenet}
Zhenqin Wu, Bharath Ramsundar, Evan~N Feinberg, Joseph Gomes, Caleb Geniesse, Aneesh~S Pappu, Karl Leswing, and Vijay Pande.
\newblock Moleculenet: a benchmark for molecular machine learning.
\newblock \emph{Chemical science}, 9\penalty0 (2):\penalty0 513--530, 2018.

\bibitem[Xu et~al.(2018)Xu, Hu, Leskovec, and Jegelka]{xu2018powerful}
Keyulu Xu, Weihua Hu, Jure Leskovec, and Stefanie Jegelka.
\newblock How powerful are graph neural networks?
\newblock \emph{arXiv preprint arXiv:1810.00826}, 2018.

\bibitem[Xu et~al.(2019)Xu, Hu, Leskovec, and Jegelka]{xu2018how}
Keyulu Xu, Weihua Hu, Jure Leskovec, and Stefanie Jegelka.
\newblock How powerful are graph neural networks?
\newblock In \emph{International Conference on Learning Representations}, 2019.
\newblock URL \url{https://openreview.net/forum?id=ryGs6iA5Km}.

\bibitem[Yang et~al.(2022)Yang, Zeng, Wu, Jia, and Yan]{yang2022learning}
Nianzu Yang, Kaipeng Zeng, Qitian Wu, Xiaosong Jia, and Junchi Yan.
\newblock Learning substructure invariance for out-of-distribution molecular representations.
\newblock \emph{Advances in Neural Information Processing Systems}, 35:\penalty0 12964--12978, 2022.

\bibitem[Yao et~al.(2022)Yao, Wang, Li, Zhang, Liang, Zou, and Finn]{yao2022improving}
Huaxiu Yao, Yu~Wang, Sai Li, Linjun Zhang, Weixin Liang, James Zou, and Chelsea Finn.
\newblock Improving out-of-distribution robustness via selective augmentation.
\newblock In \emph{International Conference on Machine Learning}, pp.\  25407--25437. PMLR, 2022.

\bibitem[You et~al.(2018)You, Ying, Ren, Hamilton, and Leskovec]{you2018graphrnn}
Jiaxuan You, Rex Ying, Xiang Ren, William Hamilton, and Jure Leskovec.
\newblock Graphrnn: Generating realistic graphs with deep auto-regressive models.
\newblock In \emph{International conference on machine learning}, pp.\  5708--5717. PMLR, 2018.

\bibitem[Yuan et~al.(2022)Yuan, Yu, Gui, and Ji]{yuan2022explainability}
Hao Yuan, Haiyang Yu, Shurui Gui, and Shuiwang Ji.
\newblock Explainability in graph neural networks: A taxonomic survey.
\newblock \emph{IEEE transactions on pattern analysis and machine intelligence}, 45\penalty0 (5):\penalty0 5782--5799, 2022.

\bibitem[Zang \& Wang(2020)Zang and Wang]{zang2020moflow}
Chengxi Zang and Fei Wang.
\newblock Moflow: an invertible flow model for generating molecular graphs.
\newblock In \emph{Proceedings of the 26th ACM SIGKDD international conference on knowledge discovery \& data mining}, pp.\  617--626, 2020.

\end{thebibliography}

\appendix
\newpage
\section{Theory}
\label{sec:proofs}

\subsection{Deriving the conditional reverse-time SDE}
\label{sec:proofs_sde}

$$
\nabla_{G_t} \log p_t\left(G_t, \mathbf{y}_G \mid \mathcal{E}_{\text {ood }}=\lambda\right)  = \nabla_{G_t} \log p_t\left(G_t\mid \mathbf{y}_G, \mathcal{E}_{\text {ood }}=\lambda\right) 
$$

Proof:

\begin{align*}
\log p_t\left(G_t, \mathbf{y}_G \mid \mathcal{E}_{\text {ood }} =\lambda\right) = & \log p_t\left(G_t, \mathbf{y}_G, \mathcal{E}_{\text {ood }}=\lambda\right) \\
- & \log p_t\left(\mathcal{E}_{\text {ood }}=\lambda\right) \nonumber
\end{align*}
Since $p_t\left(\mathcal{E}_{\text {ood }}=\lambda\right)$ is independent of $G_t$, $\nabla_{G_t} \log p_t\left(\mathcal{E}_{\text {ood }}=\lambda\right) = 0$. Therefore, 
\begin{align*}
\nabla_{G_t}\log p_t\left(G_t, \mathbf{y}_G \mid \mathcal{E}_{\text {ood }}=\lambda\right) = \nabla_{G_t}\log p_t\left(G_t, \mathbf{y}_G, \mathcal{E}_{\text {ood }}=\lambda\right)
\end{align*}
Additionally, 
\begin{align*}
\log p_t\left(G_t\mid \mathbf{y}_G, \mathcal{E}_{\text {ood }}=\lambda\right) & = \log p_t\left(G_t, \mathbf{y}_G, \mathcal{E}_{\text {ood }}=\lambda\right) \\ & - \log p_t\left(\mathbf{y}_G, \mathcal{E}_{\text {ood }}=\lambda\right)
\end{align*}
Since $p_t\left(\mathbf{y}_G, \mathcal{E}_{\text {ood }}=\lambda\right)$ is independent of $G_t$, 
$$\nabla_{G_t} \log p_t\left(\mathbf{y}_G, \mathcal{E}_{\text {ood }}=\lambda\right) = 0$$
Therefore, 
\begin{align*}
\nabla_{G_t}\log p_t\left(G_t\mid \mathbf{y}_G, \mathcal{E}_{\text {ood }}=\lambda\right) = \nabla_{G_t}\log p_t\left(G_t, \mathbf{y}_G, \mathcal{E}_{\text {ood }}=\lambda\right)
\end{align*}
Finally, 
\begin{align*}
\nabla_{G_t} \log p_t\left(G_t, \mathbf{y}_G \mid \mathcal{E}_{\text {ood }}=\lambda\right) = \nabla_{G_t} \log p_t\left(G_t\mid \mathbf{y}_G, \mathcal{E}_{\text {ood }}=\lambda\right)
\end{align*}

\section{Metrics for measuring distributional differences}
\label{sec:random_gin}
In this section, we provide detailed implementation steps for measuring the distributional differences between the augmented dataset and the training dataset. Following~\cite{thompson2022on}, we first use an untrained random GIN, $h$, to extract graph embeddings from both the augmentation distribution and the training distribution. The maximum mean discrepancy (MMD) is then computed to quantify the dissimilarity between the graph embedding distributions:
\begin{align*}
& \operatorname{MMD}^2(P \| \tilde{P}) = \mathbb{E}_{g, \tilde{g} \sim P}[k(h(g), h(\tilde{g}))]  + \\ &  \mathbb{E}_{g, \tilde{g} \sim \tilde{P}}[k(h(g), h(\tilde{g}))]  -2 \mathbb{E}_{g \sim P, \tilde{g} \sim \tilde{P}}[k(h(g), h(\tilde{g}))]
\end{align*}

where $k(\cdot, \cdot)$ is the RBF kernel proposed by~\cite{you2018graphrnn}. As recommended by~\cite{thompson2022on}, the MMD RBF scalar is also one of the most reliable metrics for measuring distributional differences:

$$
k\left(h(g), h(\tilde{g})\right)=\exp \left(-d\left(h(g), h(\tilde{g})\right) / 2 \sigma^2\right)
$$

Additionally, we employ the Earth Mover’s Distance (EMD) from~\cite{thompson2022on} to compute pairwise distances $d(\cdot, \cdot)$.

\section{Implementations of \proposedmodel. }
\label{sec:implementations_ooda}
Directly applying this framework to graphs is inadequate for addressing complex covariate shifts, especially those that simultaneously affect feature and structural distributions. To overcome this, we explicitly model the joint diffusion of node features and adjacency matrices, representing the perturbed graph as $\left\{G_t=\left(\boldsymbol{X}_t, \boldsymbol{A}_t\right)\right\}_{t=0}^T$ using a set of SDEs for Equation~\ref{eq:conditional_sde}:
\begin{equation}
\label{eq:final_reverse_sde}
\left\{\begin{aligned}
\mathrm{d} \boldsymbol{X}_t= & {\left[\mathbf{f}_{1, t}\left(\boldsymbol{X}_t\right)-(1-\sqrt{\lambda}) g_{1, t}^2 \nabla_{\boldsymbol{X}_t} \log p_t\left(\boldsymbol{X}_t, \boldsymbol{A}_t\right)\right.} \\
& \left.-(1-\sqrt{\lambda}) g_{1, t}^2 \nabla_{\boldsymbol{X}_t}\log p_t\left(\mathbf{y}_G\mid\boldsymbol{X}_t, \boldsymbol{A}_t\right)\right] \mathrm{d} \bar{t} \\
& + g_{1, t} \mathrm{~d} \overline{\mathbf{w}}_1 \\
\mathrm{~d} \boldsymbol{A}_t= & {\left[\mathbf{f}_{2, t}\left(\boldsymbol{A}_t\right)-(1-\sqrt{\lambda}) g_{2, t}^2 \nabla_{\boldsymbol{A}_t} \log p_t\left(\boldsymbol{X}_t, \boldsymbol{A}_t\right)\right.} \\
& \left.-(1-\sqrt{\lambda}) g_{2, t}^2 \nabla_{\boldsymbol{A}_t}\log p_t\left(\mathbf{y}_G\mid\boldsymbol{X}_t, \boldsymbol{A}_t\right)\right] \mathrm{d} \bar{t} \\ & + g_{2, t} \mathrm{~d} \overline{\mathbf{w}}_2 .
\end{aligned}\right.
\end{equation}
where $\mathbf{f}_t(\boldsymbol{X}, \boldsymbol{A})=\left(\mathbf{f}_{1, t}(\boldsymbol{X}), \mathbf{f}_{2, t}(\boldsymbol{A})\right)$ and $g_t=\left(g_{1, t}, g_{2, t}\right)$ representing the drift and diffusion coefficients, respectively. The reverse-time processes are captured by standard Wiener processes $\overline{\mathbf{w}}_1$ and $\overline{\mathbf{w}}_2$, with $\mathrm{d} \bar{t}$ indicating an infinitesimally small negative time step. We train one graph transformer~\cite{dwivedi2020generalization, vignac2022digress}, denoted as $s_{\theta, t}=\left(s_{\theta_1, t}, s_{\theta_2, t}\right)$ to closely estimate the partial score functions:


\[
s_{\theta,t}=\big(s_{\theta_1,t},\,s_{\theta_2,t}\big)
\approx
\Big(
\nabla_{\boldsymbol X_t}\log p_t(\boldsymbol X_t,\boldsymbol A_t),\;
\nabla_{\boldsymbol A_t}\log p_t(\boldsymbol X_t,\boldsymbol A_t)
\Big).
\]

Therefore, the denoising graph transformer is only trained with unlabelled graphs without $\lambda$ values and graph labels. The graph transformer takes perturbed node features $X_t$, adjacency matrices $A_t$ and normalized timestep as input. The timestep $t$ value is treated as a global graph feature, and an embedding layer is used to embed $t$.   

We also use a graph transformer model $\phi_t$ with the same architecture to predict the class label of the noisy graphs $G_t=\left(\boldsymbol{X}_t, \boldsymbol{A}_t\right)$ at time step $t$. The target class $j$ probability $p_t\left(\mathbf{y}_G=j\mid\boldsymbol{X}_t, \boldsymbol{A}_t\right)$ is then given by:
$$
p_t\left(\mathbf{y}_G=j\mid\boldsymbol{X}_t, \boldsymbol{A}_t\right) = \frac{e^{\phi_t\left(\boldsymbol{X}_t, \boldsymbol{A}_t\right)_{[j]}}}{\sum_{j=1}^M e^{\phi_t\left(\boldsymbol{X}_t, \boldsymbol{A}_t\right)_{[j]}}}.
$$
Once both the score transformer and the classifier are trained, we use them to compute the conditional partial scores in Equation~(\ref{eq:conditional_sde}) during the sampling process. It is noteworthy that while our generation process leverages the representational power of graph transformers, subsequent evaluations for graph OOD classification rely on augmenting only simple GNN backbones with these generated samples (as detailed in Section~\ref{secsec:experimental settings}).



We adopt the two popular time-dependent hyperparameters $\alpha_{1, t}$ and $\alpha_{2, t}$ for the target class probability predicted by $\phi_t$. These hyperparameters are defined as follows:
\begin{equation}
\begin{aligned}
\alpha_{1, t}&=0.1^t \frac{ r_{1}\left\|\boldsymbol{s}_{\theta_1, t}\left(\boldsymbol{G}_t\right)\right\|}{\left\|\nabla_{\boldsymbol{X}_t}\log p_t\left(\mathbf{y}_G\mid\boldsymbol{X}_t, \boldsymbol{A}_t\right)\right\|}, \\
\alpha_{2, t}&=0.1^t\frac{r_{2}\left\|\boldsymbol{s}_{\theta_2, t}\left(\boldsymbol{G}_t\right)\right\|}{\left\|\nabla_{\boldsymbol{A}_t}\log p_t\left(\mathbf{y}_G\mid\boldsymbol{X}_t, \boldsymbol{A}_t\right)\right\|},
\end{aligned}
\end{equation}
where $\alpha_t=\left(\alpha_{1, t}, \alpha_{2, t}\right)$, $r_{1}$ and $r_{2}$ are the weights for node features and adjacency matrices respectively, and $\|\cdot\|$ is the entry-wise matrix norm. 



Intuitively, at the early stages of the reverse-time SDEs, the graphs are highly perturbed, resembling the prior noise distribution. Therefore, the classifier cannot accurately approximate the target class probability. Consequently, in the initial denoising steps, we focus more on guiding the reverse-time SDEs towards the low-density (i.e. OOD distribution) regions. As the reverse-time SDEs progressively denoise the graphs, we introduce guidance to direct the reverse-time SDEs towards regions exhibiting the desired stable patterns and the explored OOD environmental patterns.

\section{Experimental Details}
\label{sec:experimental_details}
\subsection{Dataset Details}
We utilize six datasets from the GOOD benchmark~\cite{gui2022good}, including GOOD-Motif-base, GOOD-Motif-size, GOOD-CMNIST-color, GOOD-HIV-scaffold, GOOD-HIV-size, and GOOD-SST2-length. The GOOD benchmark~\cite{gui2022good} is the state-of-the-art framework for systematically evaluating graph OOD generalization. It carefully designs data environments to induce reliable and valid distribution shifts. The selected datasets span a diverse range of domains, covering covariate shifts in general graphs, image-transformed graphs, molecular graphs, and natural language sentiment analysis graphs. The dataset details are as follows:

\begin{itemize}
    \item \textbf{GOOD-Motif:} GOOD-Motif is a synthetic dataset from Spurious-Motif~\cite{wu2022discovering} specifically designed to investigate structure shifts. Each graph consists of an environmental base graph connected to a label-determining motif. The two primary covariate shift domains are the base graph type and graph size. For base covariate shift, the training distribution includes graphs with wheel, tree, and ladder base structures, while the validation set features star base graphs, and the test set contains path base graphs. For size covariate shift, the training distribution consists of graphs with sizes ranging from $6$ to $45$ nodes, the validation set contains graphs with sizes between $20$ and $75$ nodes, and the test set comprises graphs with sizes ranging from $68$ to $155$ nodes.
        
    \item \textbf{GOOD-CMNIST:} GOOD-CMNIST is a semi-synthetic dataset designed to investigate node feature shifts. It consists of graphs transformed from MNIST handwritten digit images using superpixel techniques~\cite{monti2017geometric}. Node color features are manually applied, making the color shift environment independent of the underlying structure. Specifically, for covariate shift, digits are colored using seven different colors. The training distribution includes digits colored with the first five colors, while the validation and test distributions contain digits with the remaining two colors, respectively.
    
    \item \textbf{GOOD-HIV:} GOOD-HIV is a small-scale, real-world molecular dataset sourced from MoleculeNet~\cite{wu2018moleculenet}. The nodes in these molecular graphs represent atoms, and the edges represent chemical bonds. This dataset is designed to study node feature shifts, edge feature shifts, and structure shifts. The two covariate shift domains are scaffold graph type and molecular graph size. For the scaffold covariate shift, environments are partitioned based on the Bemis-Murcko scaffold~\cite{bemis1996properties}, a two-dimensional structural base that does not determine a molecule's ability to inhibit HIV replication. For the size covariate shift, the training distribution consists of molecular graphs ranging in size from $17$ to $222$ atoms. The validation set contains molecules with sizes between $15$ and $16$ atoms, while the test set includes molecules with sizes from $2$ to $14$ atoms.
        
    \item \textbf{GOOD-SST2:} GOOD-SST2 is a real-world natural language sentimental analysis dataset from~\cite{yuan2022explainability}, designed to investigate node feature shifts and structure shifts. Each graph is derived from a sentence, transformed into a grammar tree, where nodes represent words, and node features are corresponding word embeddings. The task is to predict the sentiment polarity of each sentence. Sentence length is chosen as the covariate shift environment, as sentence length should not inherently affect sentiment polarity. For the length covariate shift, the training distribution consists of grammar graphs with sizes ranging from $1$ to $7$ nodes, the validation distribution includes graphs with sizes from $8$ to $14$ nodes, and the test distribution contains graphs with sizes from $15$ to $56$ nodes.
\end{itemize}

\subsection{Implementation settings}
\label{sec:implementations}
\textbf{Diffusion models:} Following~\cite{jo2022score}, we preprocess each graph into two matrices: $\boldsymbol{X} \in \mathbb{R}^{n \times a}$ for node features, and $\boldsymbol{A} \in \mathbb{R}^{n \times n \times b}$ for adjacency and edge features. Here, $n$ represents the maximum number of nodes in a graph for the given dataset, while $a$ and $b$ denote the dimensions of node features and edge features, respectively. The graph structure, including edge features, is encoded in $\boldsymbol{A}$. For the GOOD-Motif dataset, $a$ corresponds to the node degree of a node. In GOOD-CMNIST, each node feature is the concatenation of its degree and color. In GOOD-SST2, the node feature is the word embedding. In the molecular dataset GOOD-HIV, $a$ represents possible atom types and $b$ denotes the types of bonds (e.g., single, double, triple). All molecules are converted to their kekulized form, with hydrogens removed using the RDKit library~\cite{landrum2016rdkit}. Additionally, we apply the valency correction proposed by~\cite{zang2020moflow} to post-process the generated molecules.

We train a graph transformer model~\cite{dwivedi2020generalization, vignac2022digress}, $s_{\theta, t}$, to approximate the partial score functions for the unlabelled graphs in the OOD training set and evaluate them on the OOD validation set. In line with~\cite{jo2022score}, we use VP or VE SDEs to model the diffusion process for both node features and adjacency matrices. The specific details of the diffusion models are provided in Table~\ref{tab:diffusion_models_paras}.

We also train a graph transformer model, $\phi_t$, with the same architecture described in Table~\ref{tab:diffusion_models_paras}, to predict the class labels of the noisy graphs $G_t=\left(\boldsymbol{X}_t, \boldsymbol{A}_t\right)$ at each time step $t$.

\begin{table}

\caption{Hyperparameters of diffusion models.}
  \label{tab:diffusion_models_paras}  

  \begin{center}
  \resizebox{\columnwidth}{!}{
    \begin{tabular}{cccccc}
    \toprule
        & Hyperparameter & Motif & CMNIST & Molhiv & GOOD-SST2 \\
        \hline \multirow{4}{*}{$s_{\theta}$} 
         & Number of graph transformer layers & $8$ & $8$ & $9$ & $8$ \\
         & Number of attention heads & $8$ & $8$ & $8$ & $8$ \\
         & Hidden dimension of $\boldsymbol{X}$ & $256$ & $256$ & $256$ & $256$ \\
         & Hidden dimension of $\boldsymbol{A}$ & $64$ & $64$ & $64$ & $64$ \\
        \hline \multirow{4}{*}{ SDE for $\boldsymbol{X}$} & Type & VP & VP & VP & VP \\
         & Number of sampling steps & 1000 & 1000 & 1000 & 1000 \\
         & $\beta_{\text {min }}$ & 0.1 & 0.1 & 0.1 & 0.1  \\
         & $\beta_{\max }$ & 1.0 & 1.0 & 1.0 & 1.0 \\
        \hline \multirow{4}{*}{ SDE for $\boldsymbol{A}$} & Type & VP & VP & VE & VP \\
         & Number of sampling steps & 1000 & 1000 & 1000 & 1000 \\
         & $\beta_{\text {min }}$ & 0.1 & 0.1 & 0.2 & 0.2  \\
         & $\beta_{\max }$ & 1.0 & 1.0 & 1.0 & 0.8 \\
        \hline \multirow{3}{*}{ Solver } & Type & EM + Langevin & EM + Langevin & Reverse & EM \\
        & SNR & $0.2$ & $0.2$ & $0.0$ & $0.0$ \\
         & Scale coefficient & $0.7$ & $0.7$ & $0.0$ & $0.0$ \\
        \hline \multirow{6}{*}{ Train } & Optimizer & AdamW & AdamW & AdamW & AdamW \\
        & Learning rate & $4 \times 10^{-4}$ & $4 \times 10^{-4}$ & $2 \times 10^{-4}$ & $2 \times 10^{-4}$ \\
        & Weight decay & $1 \times 10^{-12}$ & $1 \times 10^{-12}$ & $1 \times 10^{-12}$ & $1 \times 10^{-12}$ \\
         & Batch size & $128$ & $64$ & $512$ & $64$ \\
        & EMA & $0.999$ & $0.999$ & $0.999$ & $0.999$ \\
        
    \bottomrule
    \end{tabular}
    }
    \end{center}

\end{table}

\textbf{Graph Out-of-Distribution Classification}: 
Following prior work~\cite{gui2022good, ligraph}, we employ GIN-Virtual~\cite{xu2018powerful, gilmer2017neural} as the GNN backbone for the GOOD-CMNIST, GOOD-HIV, and GOOD-SST2 datasets. For the GOOD-Motif dataset, we adopt GIN~\cite{xu2018how}. To ensure a fair comparison across all methods, we utilize the same GNN backbone architecture for all models.

For each experiment, we select the best checkpoints for OOD testing based on the performance on the OOD validation sets. All experiments are optimized using the Adam optimizer, with weight decay selected from $\{0, 1 \times 10^{-2}, 1 \times 10^{-3}, 1 \times 10^{-4}\}$ and a dropout rate of $0.5$. The number of convolutional layers in the GNN models is tuned from the set $\{3, 5\}$, with mean global pooling and ReLU activation. The hidden layer dimension is set to $300$. We explore the maximum number of epochs from $\{100, 200, 500\}$, the initial learning rate from $\{1 \times 10^{-3}, 3 \times 10^{-3}, 5 \times 10^{-3}, 1 \times 10^{-4}\}$, and the batch size from $\{32, 64, 128\}$. All models are trained to convergence.

For computation, we typically run each experiment on an NVIDIA GeForce RTX 4090. We report results as the mean and standard deviation across 10 random runs for all experiments.

We perform a grid search for the hyperparameter $\alpha \in \{0.1, 0.5, 1.0\}$ across all datasets. For $\lambda$, the grid search is tailored to each dataset. Specifically, we explore $\lambda \in \{0.01, 0.02, 0.03, 0.04, 0.05\}$ for the GOOD-Motif-base dataset and $\lambda \in \{0.1, 0.2, 0.3\}$ for GOOD-HIV-scaffold dataset. For GOOD-CMNIST-color, we tune $\lambda \in \{0.05, 0.1\}$. In the case of GOOD-SST2-length, where $\lambda = 0.01$ corresponds to an increase of one node in the graph size relative to the training distribution, we expand the grid search to $\lambda \in \{0.01, 0.02, ..., 0.14\}$. Similarly, for GOOD-Motif-size, where $\lambda = 0.01$ reflects an increase of one node, we use a search space of $\lambda \in \{0.01, 0.02, 0.03, 0.04, 0.05\}$. For GOOD-HIV-size, where $\lambda = 0.01$ corresponds to a decrease of ten nodes in graph size from the training distribution, we also use $\lambda \in \{0.01, 0.02, 0.03, 0.04, 0.05\}$. Since this hyperparameter tuning is performed during the sampling phase rather than the training phase, it is not computationally intensive.

\section{Baseline settings}
The implementation details for GNN backbones and hyperparameter tuning are consistent with those outlined in Appendix~\ref{sec:implementations}. For methods including ERM, IRM~\cite{arjovsky2019invariant}, GroupDRO~\cite{sagawa2019distributionally}, VREx~\cite{krueger2021out}, DANN~\cite{ganin2016domain}, Deep Coral~\cite{sun2016deep}, DIR~\cite{wu2022discovering}, DropNode~\cite{feng2020graph}, DropEdge~\cite{rong2019dropedge}, MaskFeature~\cite{thakoor2021large}, FLAG~\cite{kong2022robust}, M-Mixup~\cite{wang2021mixup}, and G-Mixup~\cite{han2022g}, we report results from the study in~\cite{ligraph}, which uses the same GNN backbones and hyperparameter tuning as specified in Appendix~\ref{sec:implementations}. For GSAT~\cite{miao2022interpretable}, CIGA~\cite{chen2022learning}, and AIA~\cite{sui2024unleashing}, we use their publicly available source code, adopting default settings and hyperparameters as detailed in their papers.

\section{Sensitivity of hyperparameter $\lambda$}
\label{sec:sensitivity_lambda}
We determine the value of $\lambda$ by evaluating its effectiveness on OOD validation sets across various datasets. The sensitivity of our method to different $\lambda$ values is illustrated in Tables \ref{tab:lambda_ablation_1}, \ref{tab:lambda_ablation_2} and \ref{tab:lambda_ablation_3}.


\begin{table}

    \caption{Performance on the GOOD-Motif-base dataset across varying $\lambda$ values.}
    \label{tab:lambda_ablation_1}   
 
    \begin{center}
    \begin{tabular}{|l|c|c|}
        \hline$\lambda$ & GOOD-Motif-base \\
        \hline 0.01 & 91.80  \\
        \hline 0.02 & 92.97  \\
        \hline 0.03 & 92.83  \\
        \hline 0.04 & 93.03  \\
        \hline 0.05 & 92.77  \\
        \hline
    \end{tabular}
    \end{center}


\end{table}

\begin{table}
    \caption{Performance on the GOOD-HIV-scaffold dataset across varying $\lambda$ values.}
    \label{tab:lambda_ablation_2}

    \begin{center}
    \begin{tabular}{|c|c|}
        \hline$\lambda$ & GOOD-HIV-scaffold \\
        \hline 0.1 & 78.94 \\
        \hline 0.2 & 77.73 \\
        \hline 0.3 & 77.42 \\
        \hline
    \end{tabular}
    \end{center}

\end{table}

\begin{table}
    \caption{Performance on the GOOD-CMNIST-color dataset across varying $\lambda$ values.}
    \label{tab:lambda_ablation_3}
    
    \begin{center}
    \begin{tabular}{|c|c|}
        \hline$\lambda$ & GOOD-CMNIST-color \\
        \hline 0.05 & 68.66 \\
        \hline 0.1 & 67.91 \\
        \hline
    \end{tabular}
    \end{center}

\end{table}




\section{Statistical Significance}
\label{sec:t_test}
Statistical significance is evaluated via a paired nonparametric bootstrap test~\cite{davison1997bootstrap} at the $5 \%$ level. The observed gains are significant across settings: on GOOD-Motif (base), \proposedmodel{} improves over DropNode by $0.70 \%$ with $p=0.0305$; on GOOD-Motif (size), over AIA by $4.96 \%$ with $p=$ 0.0015 ; on GOOD-CMNIST (color), over AIA by $18.23 \%$ with $p=4.9998 \times 10^{-5}$; on GOOD-HIV (scaffold), over DropNode by $1.49 \%$ with $p=0.0147$; on GOOD-HIV (size), over DANN by $1.28 \%$ with $p=0.0062$; and on GOOD-SST2 (length), over MaskFeature by $0.69 \%$ with $p=0.0207$.

\section{Time and memory complexity}
\label{sec:time_complex}
Our pipeline consists of two stages: data augmentation, and trainng of the GNN classifier on augmented graphs. The second stage of classification follows general GNN training setup, without introducing additional complexity: the time/memory complexity per layer of using Graph Isomorphism Networks (GIN) as backbone is $\Theta(n + e)$, where $n$ is the number of nodes and $e$ is the number of edges. For the data augmentation stage, we introduce graph transformer whose memory and time complexity per layer is $\Theta(n^2)$. This arises from the computation of attention scores and predictions across each edge.

\end{document}